\documentclass{article} 
\usepackage{iclr2025_conference,times}

\usepackage{subcaption} 
\usepackage{adjustbox}
\usepackage{multirow}
\usepackage[table]{xcolor} 

\usepackage{listings}

\usepackage[ruled,vlined,linesnumbered]{algorithm2e}

\usepackage{enumitem}

\usepackage{xspace}
\usepackage{amsmath}
\usepackage{booktabs}
\usepackage{wrapfig}
\usepackage{caption}
\usepackage[utf8]{inputenc}
\usepackage{tcolorbox}
\tcbuselibrary{skins}
\tcbuselibrary{breakable}
\usepackage{enumitem}
\usepackage{float}
\usepackage{hyperref} 

\newcommand{\header}[1]{\smallskip \noindent\textbf{{#1}}\xspace}

\newcommand{\papertoweb}{\textsc{Paper2Web}\xspace}
\newcommand{\dataset}{\textsc{Paper2Web}\xspace}
\newcommand{\agent}{\textsc{PWAgent}\xspace}
\newcommand{\WebsiteAgent}{\textsc{PWAgent}\xspace}
\newcommand{\Description}[1]{}

\newcommand{\redref}[1]{\textcolor{red}{\ref{#1}}}

\usepackage{titletoc}  
\usepackage{titlesec}  
\usepackage{url}
\usepackage{graphicx}
\usepackage{wrapfig}        
\usepackage[font=small,labelfont=bf]{caption} 
\usepackage{float}          
\usepackage{fancyvrb}
\usepackage{lipsum}         
\usepackage{minted}
\usepackage{listings}
\usepackage{wrapfig}
\usepackage{fontawesome5}
\title{\papertoweb: Let's Make Your Paper Alive!}

\author{Yuhang Chen$^{1*}$,
Tianpeng Lv$^{1*}$,
Siyi Zhang$^{1}$,
Yixiang Yin$^{1}$,
Yao Wan$^{1\dagger}$,\\
\textbf{Philip S. Yu}$^{2}$, \textbf{Dongping Chen}$^{3\ddagger}$ \\
$^{1}$ONE Lab, Huazhong University of Science and Technology, \\
$^{2}$University of Illinois Chicago, $^{3}$University of Maryland \\
\texttt{\{u202315752, wanyao\}@hust.edu.cn}, \texttt{dongping@umd.edu} \\
\footnotesize{$^{*}$ Equal Contribution. $^{\dagger}$ Corresponding author. $^{\ddagger}$Project Lead.}
}

\iclrfinalcopy 
\begin{document}

\maketitle

\begin{center}
    \vspace{-2em}
    {\large \faGlobe~Project Website:}
    \textbf{ \url{https://francischen3.github.io/P2W_Website}}
    
    {\large \faGithub~Code Repository:}
    \textbf{ \url{https://github.com/YuhangChen1/Paper2All}}
\end{center}

\begin{figure}[!h]
  \centering

  \includegraphics[scale=0.55]{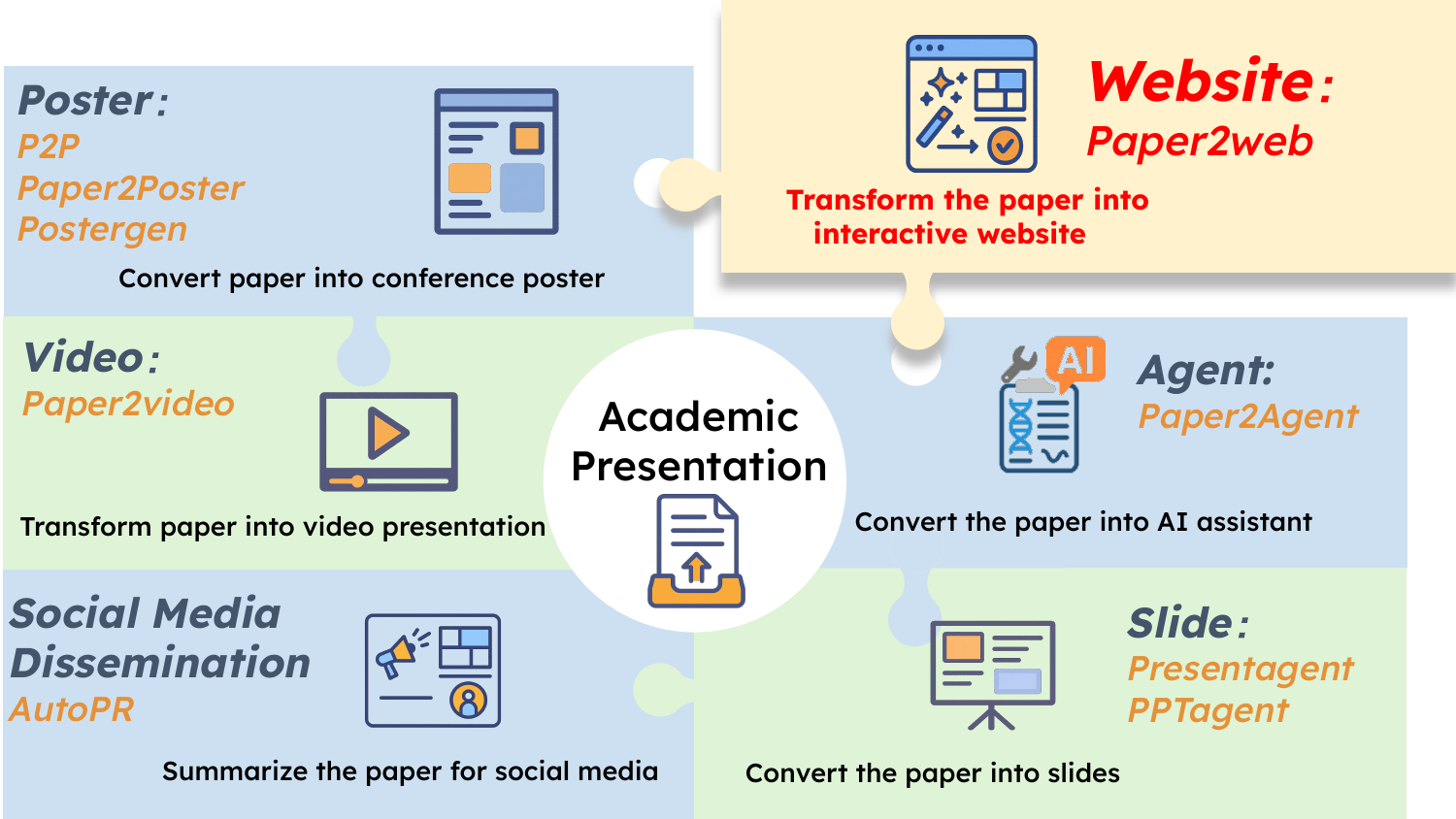}
  \caption{Our work, \papertoweb, constitutes an important piece of the puzzle for the presentation and dissemination of academic papers. We build a unified platform to streamline all academic presentation  at \href{https://github.com/YuhangChen1/Paper2All}{Paper2All}.}
  \Description{circle}
  \label{circle}
\end{figure}

\begin{abstract}
Academic project websites can more effectively disseminate research when they clearly present core content and enable intuitive navigation and interaction. However, current approaches such as direct \textit{Large Language Model} (LLM) generation, templates, or direct HTML conversion struggle to produce layout-aware, interactive sites, and a comprehensive evaluation suite for this task has been lacking.
In this paper, we introduce \papertoweb, a benchmark dataset and multi-dimensional evaluation framework for assessing academic webpage generation. It incorporates rule-based metrics like \textit{Connectivity}, \textit{Completeness} and human-verified LLM-as-a-Judge (covering interactivity, aesthetics, and informativeness), and PaperQuiz, which measures paper-level knowledge retention. We further present \WebsiteAgent, an autonomous pipeline that converts scientific papers into interactive and multimedia-rich academic homepages. The agent iteratively refines both content and layout through MCP tools that enhance emphasis, balance, and presentation quality.
Our experiments show that \WebsiteAgent consistently outperforms end-to-end baselines like template-based webpages and arXiv/alphaXiv versions by a large margin while maintaining low cost, achieving the Pareto-front in academic webpage generation.

\end{abstract}

\section{Introduction}
Research papers are predominantly distributed in PDF format, conveying information solely through static text and images~\citep{tkaczyk2015cermine,li2020docbank,clark2016pdffigures2,lo2020s2orc}. However, PDFs offer limited support for interactivity and multimedia content~\citep{w3c_reflow,govuk_accessible_documents,nhs_avoid_pdfs,kumar2024pdfaccessibility}, resulting in substantial information loss during dissemination~\citep{tkaczyk2015cermine,li2020docbank}. As a result, transforming academic papers into more visual and accessible formats has emerged as a promising direction for enhancing scholarly communication and accelerating knowledge dissemination~\citep{fischhoff2013sciences,thorlacius2007role}.

Recently, growing efforts have sought richer and more efficient ways to transform scholarly articles—such as converting papers into concise posters with Paper2Poster~\citep{pang2025paper2poster}, presentation slides with PresentAgent~\citep{shi2025presentagent}, videos with Paper2Video~\citep{paper2video}, public-facing content with AutoPR~\citep{chen2025autopr}.
However, these approaches either discard the fine-grained details present in the original text or retain only the main ideas while overlooking the communicative advantages of multimedia content such as videos and animated graphics. This creates a gap for formats that preserve core textual knowledge while seamlessly integrating multimedia to enhance scientific communication across diverse communities. 

\begin{figure}[!b]
  \centering
  \begin{subfigure}[t]{0.5\linewidth}
    \centering
    \includegraphics[width=\linewidth]{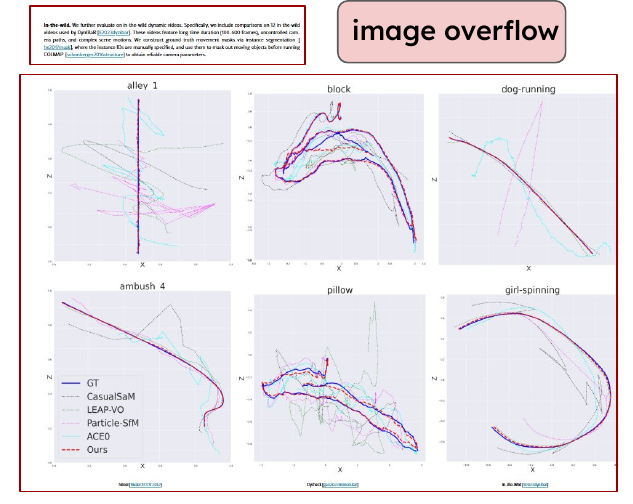}
    \caption{Web page of arXiv HTML version.}
    \label{fig:2a}
  \end{subfigure}
  \hfill
  \begin{subfigure}[t]{0.45\linewidth}
    \centering
    \includegraphics[width=\linewidth]{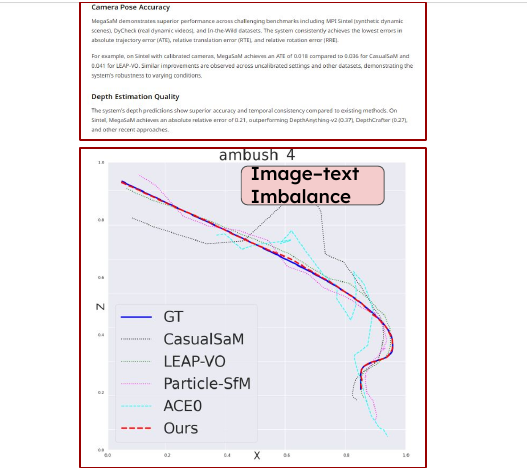}
    \caption{Web page generated by alphaXiv.}
    \label{fig:2b}
  \end{subfigure}

  \caption{Problems in current scholar web page generation, including distorted layout and limited interactivity.}
  \label{fig:web_comparison}
\end{figure}

Compared with the above methods, an online web page can integrate textual content with multimedia and present in a coordinated and navigable manner. As illustrated in Figure~\redref{datax}, well-designed webpages can bridge the gap between scholarly content and interactive digital presentation, thereby enabling broader and more effective dissemination of research outcomes. However, this poses challenges in requiring deliberate spatial organization to accommodate rich media and interactive components. Recent efforts have explored converting full academic papers into web pages to broaden accessibility and dissemination. The arXiv HTML initiative~\citep{frankston2024html} is one representative example, yet such approaches often produce disordered layouts and redundant text, reducing readability, precision, and cross-device accessibility. As illustrated in Figure~\redref{fig:web_comparison}, common failure modes include rigid figure grids with inconsistent scaling, detached captions, missing responsiveness, and limited interactivity. AlphaXiv leverages LLMs for content condensation and layout optimization, yet it still limits author control over multimedia placement and visual design, resulting in largely static presentations that fail to fully exploit interactive capabilities. As noted by prior work~\citep{frankston2024html}, these issues stem from TeX–HTML pipelines that emulate LaTeX behavior without executing a full TeX engine, leading to missing structures and visual inconsistencies. On the other hand, directly LLM-driven webpage generation also struggles to process long contexts~\citep{liu-etal-2024-lost,hsieh2024ruler} and to effectively integrate multimedia content while maintaining robust interactivity~\citep{xiao2024interaction2code}. 

\begin{wrapfigure}{r}{0.5\textwidth}
    \centering
    \includegraphics[width=1.05\linewidth]{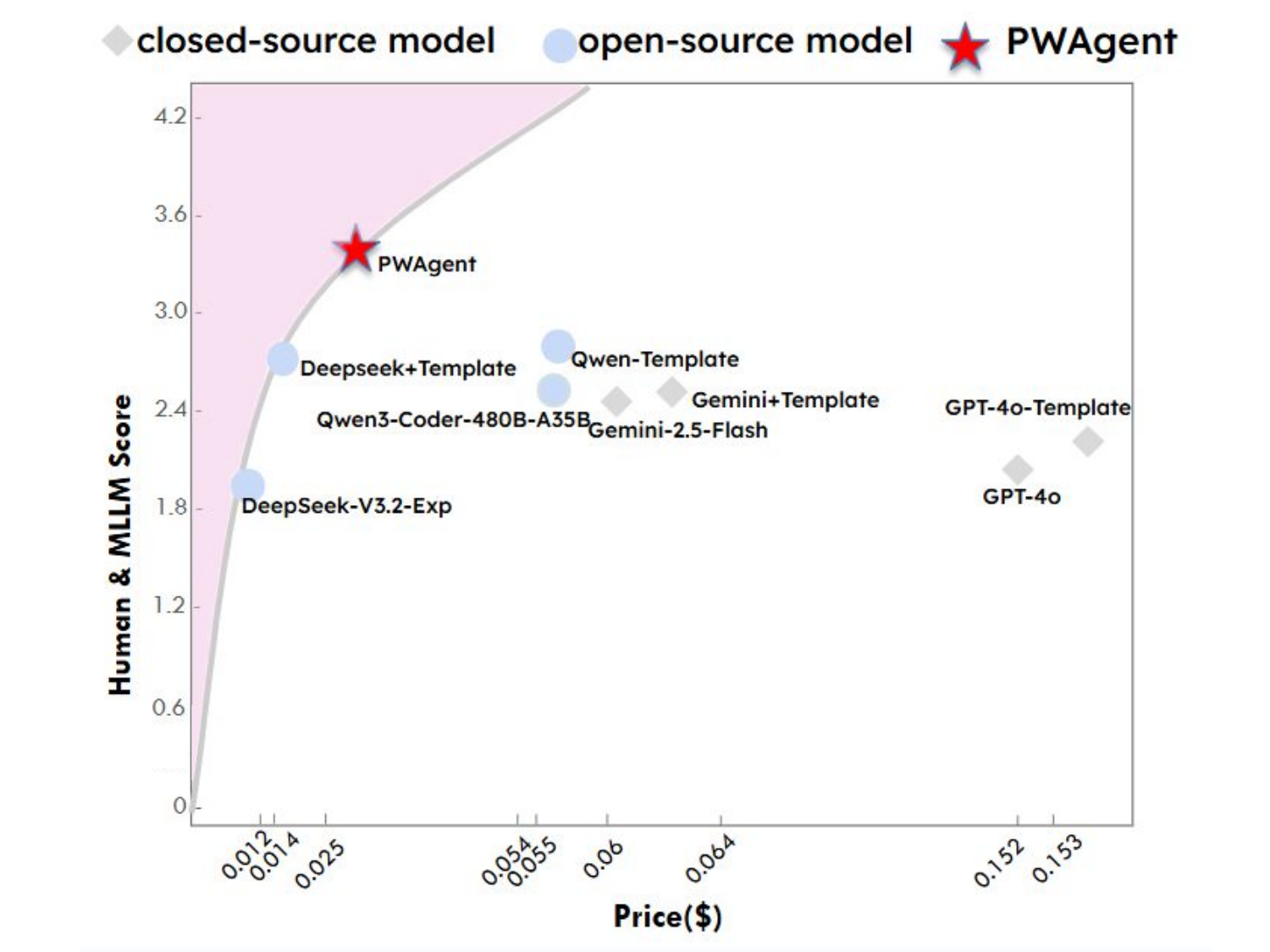}
    \vspace{-2em}
    \caption{Pareto-front comparison of each website generation methods. Our \agent achieve the highest quality with moderate and affordable cost.}
    \label{fig:pareto}
    \vspace{-1em}
\end{wrapfigure}
 
\header{Our Work: \papertoweb.}
In this paper, we introduce \papertoweb, a new task that aims to transform full academic papers into interactive websites that preserve core content while integrating multimedia and improving usability. We begin by constructing a dataset of paired academic papers and their corresponding webpages. Specifically, we crawl accepted papers from selected conferences, parse their texts to extract reliable metadata such as authorship, and augment each record with citation counts from Semantic Scholar. We then perform multi-stage filtering: using cues in the paper body and related code repositories to locate candidate project homepages, employing an LLM to assess their relevance, and relying on human annotators to resolve ambiguous cases. This pipeline yields 10,700 papers with verified homepages, forming the basis for our analysis of effective research websites.
To address these limitations, we propose \WebsiteAgent, a multi-agent framework for transforming scholarly documents into structured and interactive web content. \WebsiteAgent first decompose a paper into structured assets and organize links and executable artifacts under a unified schema. It then performs \textit{Model Context Protocol} (MCP) ingestion to construct a semantically aligned resource repository enriched with relational metadata and exposed through standardized tools for downstream use. A content-aware allocation heuristic estimates each asset’s spatial footprint and assigns provisional layout budgets to guide rendering and navigation. Finally, agent-driven iterative refinement drafts an initial website, inspects rendered views, and issues targeted edits via tool calls to correct visual imbalance, enhance information hierarchy, and appropriately anchor multimedia elements. This loop alternates between global assessment and localized adjustment, linking segmented screenshots to corresponding HTML fragments for precise editing.

Using the \dataset dataset, we also construct a benchmark for \papertoweb. We introduce the first metric to measure the interactivity and dynamic elements of the webpage, as well as \textit{Connectivity} and \textit{Completeness}, human-assisted \textit{MLLM-as-a-Judge} for comprehensive assessments.  Furthermore, we propose PaperQuiz to evaluate knowledge transfer from webpage screenshots through both verbatim and interpretive questions, incorporating a verbosity penalty to discourage overly text-heavy pages.
On this benchmark, \WebsiteAgent improves connectivity and completeness by roughly 12\% on average across methods, achieving a 28\% gain over the arXiv HTML baseline. It also yields an 18\% average improvement via MLLM-as-a-Judge and \textbf{triples} the average score of the strongest end-to-end baseline and remains competitive with template-assisted variants.

\header{Contributions.}
The key contributions of this paper are as follows:

\begin{itemize}[leftmargin=*]
\item \textbf{A New Task, Dataset and Evaluation Suite}. We build the \papertoweb dataset, a large-scale corpus that links scientific papers to their corresponding project homepages, enabling quantitative analysis of web-based academic dissemination.
\item \textbf{Comprehensive Benchmark}. We establish a benchmark with autonomous metrics aligned well with human preference to comprehensively assess the quality of web page generation, reveal problems within current automatic webpage generation methods.
\item \textbf{A \emph{State-of-the-Art} Automatic Approach}. We propose \WebsiteAgent, a MCP-based agent for the end-to-end transformation of academic papers into structured,interactive pages.

\end{itemize}

\section{\dataset: A New Task and Dataset}
\begin{figure*}[!t]
\centering
\includegraphics[width=0.8\textwidth]{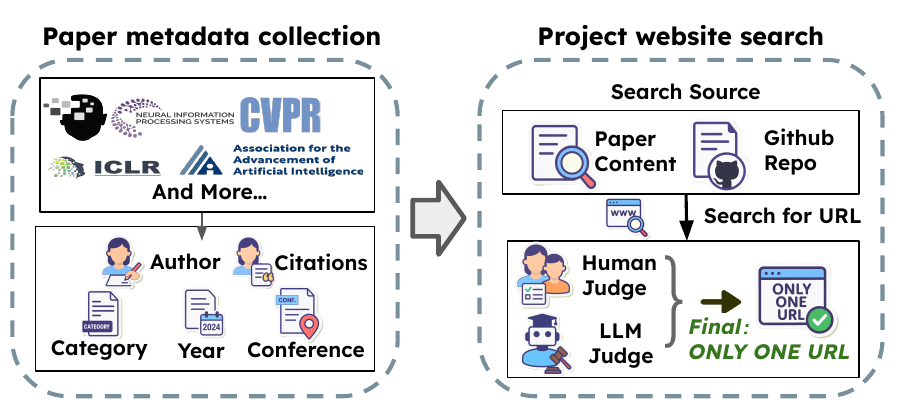}
\vspace{-1em}
\caption{To transform static papers into exploratory web pages, we collect the first paper-webpage dataset by crawling across multiple top-tier conferences and filtering by online search and human annotators.}
\label{2_5}
\vspace{-1em}
\end{figure*}

Since no dataset exists for analyzing academic website content and layout, we collect data from recent AI papers. We harvest project links from papers and code repositories, then crawl the corresponding webpage. Finally, we collect a comprehensive dataset covering multiple conferences and categories with 10,716 papers and their human-created project homepages.
Figure \redref{2_5} presents our data collection pipeline.

\subsection{Data Collection}

\begin{figure*}[!h] 
  \centering

  \includegraphics[scale=1.0]{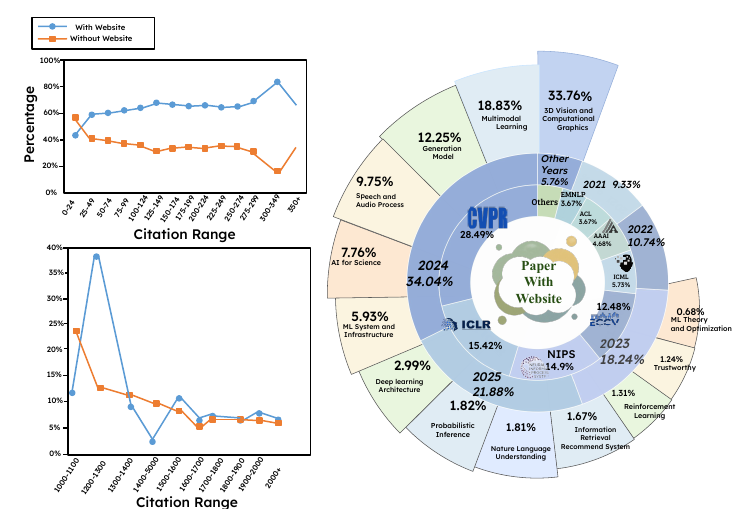}
  \caption{The right panel shows the categorization of our data. We divided the dataset into 13 categories and counted items in each. In addition, we show distributions by conference and by year. The top-left panel presents, for each category, the relative proportions of papers without and with a website among papers with low citation counts.
The bottom-left panel depicts the distribution of papers without and with a website restricted to highly cited papers (those with over 1,000 citations).}
  \Description{datax}
  \label{datax}
\end{figure*}

\header{Paper Metadata Collection.}
We focus on AI papers as they are recent, peer-reviewed, cover diverse subfields with varied modalities, and attract attention that motivates high-quality dissemination. Using automated tools, we collect papers from major AI conferences (ICML, NeurIPS, WWW, ICLR, etc., 2020-2025). We extract source links, parse full texts for metadata (title, authors, venue, year), and retrieve citation counts from Semantic Scholar. Each paper's introduction is submitted to an LLM that assigns one of thirteen topical categories (Figure~\redref{datax}, right panel), enabling standardized cross-paper analysis.

\begin{wrapfigure}{r}{0.3\textwidth}
    \centering
    \vspace{-2em}
    \includegraphics[width=\linewidth]{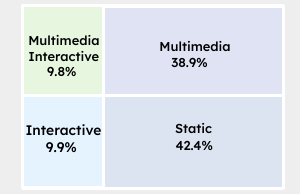}
    \caption{\dataset dataset statistics.}
    \label{fig:2c}
    \vspace{-3em}
\end{wrapfigure}
\header{Project Website Search.}
Our pipeline retrieves external links from each paper and its code repository, scanning the paper body and README files. We parse local context around each link, crawl the target HTML, and use an LLM to analyze the content. Human reviewers resolve ambiguous cases to ensure each paper maps to at most one canonical project website. Papers lacking relevant links in either source are defined as having no project homepage.

\subsection{Data Characteristics}
Finally, we curate a comprehensive dataset comprising 10,716 papers with human-created project homepages and 85,843 without. We group papers into 13 categories following ICML/NeurIPS/ICLR conference taxonomies. The right panel of Figure~\redref{datax} shows computer vision has the strongest demand for project websites, with homepage adoption rising steadily in recent years. To characterize webpage features, we manually audited 2,000 samples. We define interactive sites as pages with dynamic behaviors and explorable components responding to user intent; multimedia pages as those embedding rich media like videos; and static sites as pages delivering primarily text and still images in linear presentation. Figure~\redref{fig:2c} shows the distribution by feature set. While many pages remain static, multimedia dissemination through embedded videos and animations is notable. Interactive capabilities that enhance user experience remain comparatively rare and unevenly implemented, representing the first systematic characterization of interactive behavior and multimedia orchestration in academic webpages.

\section{Evaluation Metrics}\label{metric}

To systematically assess the quality of generated academic web pages, we introduce the \papertoweb benchmark. This comprehensive metric suite centers on the dual principles of information efficiency and a balanced text–visual composition. The framework evaluates web pages across three key dimensions: (1) Connectivity \& Completeness, (2) Holistic Evaluation via Human/MLLM-as-a-Judge, and (3) PaperQuiz, which measures how effectively the website transfers knowledge.

\subsection{Connectivity \& Completeness}
This metric jointly evaluates the hyperlink quality and structural fidelity of generated web pages. Both indicators are assessed through LLM analysis of the HTML source code, supplemented by human evaluation for reliability. For connectivity, we examine how effectively the webpage links internal and external resources to support coherent navigation and information access.
To reduce evaluation bias, a dedicated URL parser is employed to count and verify valid hyperlinks, ensuring objective measurement of link quality. For completeness, we measure how well the generated webpage reproduces the core sections of the source paper. To enhance consistency, two quantitative priors, image–text balance and information efficiency, are applied to further evaluate structural integrity and content compactness.

\header{Image–Text Balance Prior.}
Let $D$ denote the weighted deviation between the observed image–text ratio and the ideal $1{:}1$ balance, and let $\gamma>0$ be a scaling factor~\citep{pang2025paper2poster}.
We define the penalty term and score as:
\begin{equation}\label{eq1}
\zeta = \frac{5}{1+\gamma\cdot D}\,, \qquad S_{\text{img-txt}} = 5 - \zeta\,.
\end{equation}
\header{Information Efficiency Prior.}
To encourage concise, information-dense presentation, let $r=L/W$ denote the ratio between the generated text length $L$ and the median human-designed length $W$, with $\beta>0$ a scaling factor (e.g., $\beta{=}0.6$)~\citep{tufte1983visual}.
We define the efficiency as:
\begin{equation}\label{eq:info_prior}
p(r) = \frac{5}{1+\beta\cdot\max(0,r-1)}\,.
\end{equation}

\subsection{Holistic Evaluation with Human-Verified MLLM-as-a-Judge}
To evaluate the overall effectiveness of web pages at a holistic level, we employ a MLLM as an automated judge, combined with human verification to mitigate bias. The model outputs a quantitative score ranging from 1 to 5 for each webpage. Specifically, it evaluates three key dimensions: \textit{Interactive}, which measures element responsiveness, saliency emphasis, and overall usability; \textit{Aesthetic}, which assesses element quality, layout balance, and visual appeal; and \textit{Informative}, which evaluates the clarity and logical coherence of webpage content. See Appendix \ref{Appendix:human} for scoring guidelines.

\subsection{PaperQuiz}
Inspired by Paper2Poster~\citep{pang2025paper2poster}, we focus on the academic web page and acknowledge its central role in communicating research as a dynamic bridge between authors and a broader audience. Therefore, we design an evaluation protocol that simulates this knowledge-transfer scenario. We first employ an LLM as an examiner to generate a comprehensive set of 50 questions from the source paper. These questions are divided into two types: 25 Verbatim questions, which are directly answerable from specific text, figures, or tables on the webpage, and 25 Interpretive questions, which require a higher-level comprehension of the paper's core contributions, methodology, and results. In the second stage, we present a screenshot of the rendered webpage to a diverse panel of MLLMs (including both open and closed source models). These models are tasked with answering the quiz based solely on the provided webpage content. By comparing the quiz scores across different generated web pages, we can quantitatively assess which one most effectively conveys the original paper's essential information. To prevent high scores resulting from excessive text transfer, we introduce a penalty term $\zeta$, defined in Eq.~\redref{eq1}, to discount for verbosity.

\begin{figure*}[!t]
\centering
\includegraphics[width=0.85\textwidth]{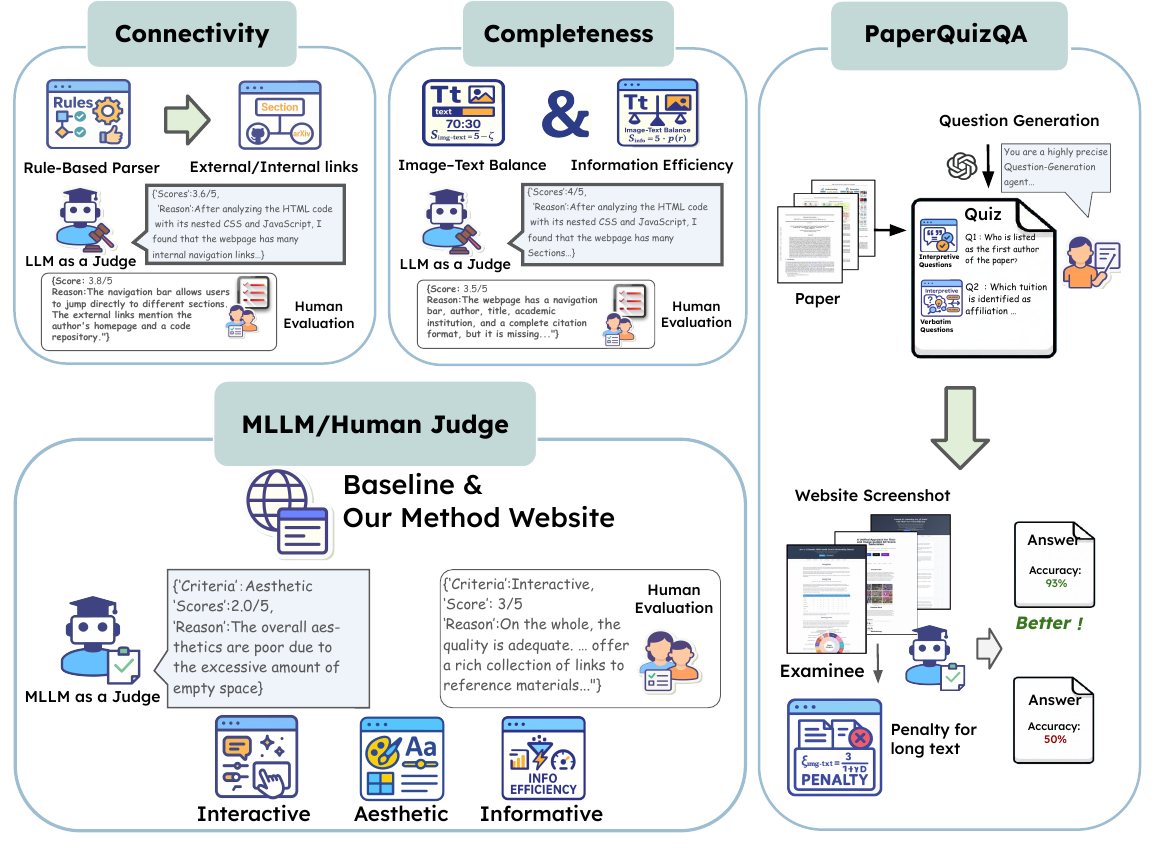}
\vspace{-1em}
\caption{Our evaluation metrics include multiple modules: (1) \textit{Connectivity} and \textit{Completeness} by parsing HTML links and structure with image–text balance and information-efficiency priors, (2) an MLLM/Human Judge to rate interactivity, aesthetics, and informativeness in a holistic manner, and (3) a QA PaperQuiz on webpage screenshots with a verbosity penalty.}
\label{3_5}
\vspace{-1em}
\end{figure*}
\section{\agent: A Strong Baseline}

To address the core challenges of the \papertoweb, we introduce \WebsiteAgent, an automated pipeline for converting scientific papers into project homepages. The core of our approach involves parsing the paper's content into a structured format managed by an MCP server~\citep{hou2025model,ehtesham2025survey,krishnan2025advancing}. This server encapsulates key paper assets, along with predefined prompts for webpage generation and stylistic refinement, organizing them into a centralized resource repository. During this process, the agent leverages the tool-use capabilities of the MCP to access the resource repository, enabling a continuous optimization loop. The overall process includes the following key stages: (1) \textit{Paper Decomposition}, which isolates key contributions from the paper. (2) \textit{MCP Ingestion}, which encapsulates these contributions as a resource repository managed by the MCP server. (3) \textit{Agent-driven Iterative Refinement}, which connects the MCP server to LLM-based agents that autonomously perform content matching and optimization through tool calls.

\subsection{Paper Decomposition}\label{first-stage}
We first deconstruct an unstructured scientific paper into structured intellectual assets that populate the MCP Resource Repository. Starting from the source PDF, the document is converted to Markdown using tools such as MARKER\footnote{https://github.com/datalab-to/marker} or DOCLING\footnote{https://github.com/docling-project/docling}. An LLM then performs semantic decomposition that extracts metadata, reconstruct tables, and model detailed page layout and reading order, yielding a machine-readable representation like JSON or Markdown that captures the paper’s key contributions.

Instead of summarizing, the LLM analyzes the Markdown text against a predefined schema to identify, isolate, and organize the paper’s key assets. These assets fall into three categories: (1) \textit{Textual Assets:} each logical section is represented as a distinct resource object containing its title, LLM-generated synopsis, full text, and metadata; (2) \textit{Visual Assets:} figures and tables are extracted as images and linked to their original captions, labels, and textual references to preserve context; and (3) \textit{Link Assets:} external URLs and internal citations are systematically captured and categorized to provide structured access to supplementary materials and related work.

\begin{figure*}[!t]
\includegraphics[width=\textwidth]{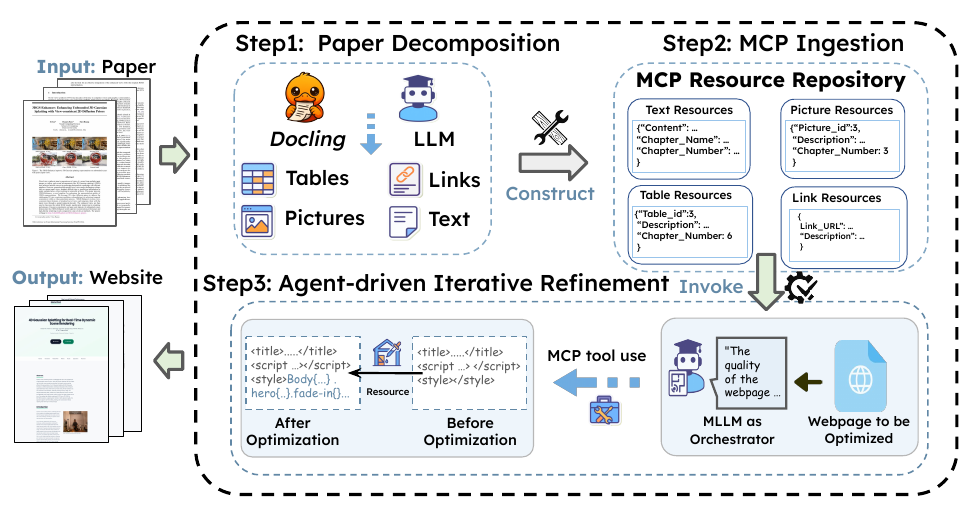}

\caption{\WebsiteAgent turns papers into interactive and multimedia-rich project homepages. Papers are deconstructed via Docling/Marker + LLM into multiple assets and stored in an MCP repository. An agent drafts a page, then iteratively optimizes until layout and UX are solid.}
\Description{Agent}
\label{3}
\vspace{-1em}
\end{figure*}

\subsection{MCP Ingestion}
Here we apply the MCP to the task of transforming scholarly papers into structured, queryable resources. We first instantiate a fully instrumented MCP server, which converts static assets into queryable resources with stable IDs and standardized tool access points. The server is responsible for resource construction, materializing assets with relational metadata and provisional layout budgets, and for tool registration, exposing a minimal, consistent API for downstream retrieval, composition, and editing.

We enrich the parsed outputs with cross-modal semantics: \textbf{(1)} An LLM is used to align each visual element with its most relevant textual description and adds back-references to the citing paragraphs. \textbf{(2)} Link assets are typed by function to support structured cross-references. To achieve a coherent visual presentation, a content-aware spatial allocation heuristic estimates each asset’s footprint and assigns a proportional layout budget to balance visual density across the page.

These enriched records are then committed to MCP server as MCP Resource Repository, where each resource is stored with a unique rid and fields for grounding and navigation. Concretely, the text resource stores the full paragraph and an LLM-generated synopsis; a Visual resource stores the image and its caption; and a Link resource stores the URL, its semantic role, and a short descriptor. Together, these resources form a structured, cross-referenced repository that serves as the foundation for webpage synthesis.
Finally, the server registers a compact tool suite that provides enumeration of resource IDs, access to grounded content and metadata for rendering, typed references for connectivity-aware placement, and initial layout allocation. This lightweight yet expressive interface is sufficient to synthesize a well-grounded HTML first draft for subsequent refinement by the multi-agent workflow.


\subsection{Agent-Driven Iterative Refinement}
Finally, we propose an agent-driven iterative refinement mechanism to progressively enhance the layout, visual coherence, and semantic alignment of generated webpages. The process begins with initial page generation, where the agent retrieves essential metadata and relevant assets from the resource repository using MCP tools. Based on this information, it rapidly constructs a foundational webpage that serves as the baseline for subsequent refinement.

Following initialization, the system enters an iterative refinement loop that continues until no further corrective actions are needed or a predefined iteration limit is reached. At its core is an MLLM acting as the \textit{Orchestrator Agent}, which conducts holistic visual assessments of the rendered webpage and invokes MCP tools to fix detected flaws. To address complex layout and visual consistency issues, the Orchestrator performs joint global–local reasoning and coordinates targeted optimizations through tool calls. To reduce hallucinations during long-range reasoning, the agent segments the rendered page into independent visual tiles linked to their corresponding HTML fragments, sequentially analyzing each to detect imbalances and misalignments and propose precise edits. After each round of local refinement, adjacent tiles are merged, borrowing the spirit of merge sort. Therefore, neighboring regions can be jointly optimized by integrating their HTML and imagery. This aggregation allows the MLLM to capture inter-section dependencies and prevent visual artifacts such as overflow, occlusion, or cross-section drift. Finally, the Orchestrator performs a global pass to assess overall content completeness and visual harmony, realizing a part-to-whole optimization path that further mitigates hallucinations. The process terminates once optimization is complete or the maximum refinement cycles are reached.

\section{How \agent Make Paper Alive?}

\subsection{Experiment Setups}
We evaluate four distinct baseline methodologies to rigorously assess the performance of our proposed approach. These serve as crucial benchmarks for gauging information dissemination efficacy and human-centered friendliness. \textbf{(1) Oracle Method}, original websites created by authors. They serve as the gold standard for optimal presentation and content delivery; \textbf{(2) End-to-End Generation}, where GPT-4o, \textit{Gemini-2.5-Flash} (Gemini), \textit{DeepSeek-V3.2-Exp} (DeepSeek) and \textit{Qwen3-Coder-480B-A35B} (Qwen) generate websites either through text-based rendering from scratch or by adapting the widely adopted Nerfies academic website template~\citep{park2021nerfies} 
(The above models combined with a template will be referred to respectively as GPT-4o-Template, Gemini-Template, DeepSeek-Template, and Qwen-Template); 
\textbf{(3) Existing HTML Versions}, where research papers from arXiv and alphaXiv provide public HTML versions, we scrape their screenshots and source code, noting some lack official web formats; \textbf{(4) \WebsiteAgent (Our)}, where Qwen3-30B-A3B is responsible for paper deconstruction and MCP ingestion, while the Orchestrator Agent is powered by the Qwen2.5-VL-32B model.

\begin{table*}[!t]
\caption{Detailed comparison between \papertoweb and other baselines across \textit{Completeness}, \textit{Connectivity} and holistic MLLM evaluation.}  
\label{tab1}
\setlength{\tabcolsep}{2pt}
\centering
\begin{adjustbox}{width=\textwidth,center}
\begin{tabular}{l c c c c | c c c c | c c c | c c c | c c c}
\toprule
\textbf{Methods} & \multicolumn{4}{c|}{\multirow{3}{*}{\textbf{Connectiveness}}} & \multicolumn{4}{c|}{\multirow{3}{*}{\textbf{Completeness}}} & \multicolumn{9}{c}{\textbf{Holistic Evaluation}} \\
 \cmidrule(lr){10-18}
& \multicolumn{4}{c|}{} & \multicolumn{4}{c|}{} & \multicolumn{3}{c|}{\textbf{Interactive}} & \multicolumn{3}{c|}{\textbf{Aesthetic}} & \multicolumn{3}{c}{\textbf{Informative}} \\
\cmidrule(lr){10-12} \cmidrule(lr){13-15} \cmidrule(lr){16-18}
& Rule $\uparrow$ & LLM  $\uparrow$ & Human $\uparrow$  & Avg.$\uparrow$ & Rule$\uparrow$ & LLM$\uparrow$ & Human$\uparrow$ & Avg.$\uparrow$ & MLLM$\uparrow$ & Human$\uparrow$ & Avg.$\uparrow$ & MLLM$\uparrow$ & Human$\uparrow$ & Avg.$\uparrow$ & MLLM$\uparrow$ & Human$\uparrow$ & Avg.$\uparrow$ \\
\midrule
\rowcolor{gray!20}
Original Website & 3.20 & 3.47 & 2.99 & $3.22$ & 3.17 & 3.93 & 4.00 & $3.70$  & 1.70 & 3.37 & $2.54$  & 3.14 & 3.63 & $3.39$ & 4.49 & 3.86 & $4.18$ \\
\midrule
\multicolumn{18}{l}{\textit{Model end-to-end methods }}  \\
GPT-4o & 1.81 & 2.07 & 2.05 & $1.98$ & 2.11 & 3.15 & 3.43 & $2.90$ & 0.53 & 1.85 & $1.19$ & 2.61 & 2.13 & $2.37$ & 2.01 & 2.56 & $2.29$ \\
Gemini-2.5-flash & 2.26 & 2.11 & 2.16 & $2.18$ &  2.72 & 3.56 & 3.43 & $3.24$ & 1.30 & 2.15 & $1.73$ & \underline{2.80} & 2.41 & $2.61$ & 3.63 & 2.68 & $3.16$ \\
DeepSeek-V3.2-Exp & 1.83 & 2.09 & 2.16 & $2.03$ & 2.09 & 3.21 & 3.51 & $2.94$ & 0.54 & 2.01 & $1.28$ & 2.63 & 2.20 & $2.42$ & 2.21 & 2.61 & $2.41$ \\
Qwen3-Coder-480B-A35B & 2.52 & 3.05 & 2.82 & $2.80$ &  2.79 & 3.58 & 3.62 & $3.33$ & 1.44 & \underline{2.43} & $\,\underline{1.94}$ & 2.74 & 2.49 & $2.62$ & 3.92 & 2.81 & $3.37$ \\
\midrule
\multicolumn{18}{l}{\textit{Model end-to-end methods \textbf{+} Template}}  \\
GPT-4o-Template & 1.83 & 2.26 & 2.77 & $2.29$ & 2.25 & 3.37 & 3.54 & $3.05$ & 0.56 & 1.47 & $1.02$ & 2.63 & 2.35 & $2.49$ & 3.87 & 2.58 & $3.23$ \\
Gemini-Template & 2.47 & 2.87 & 2.78 & $2.71$ & 2.73 & 3.72 & 3.78 & $3.41$ & \textbf{1.47} & 1.58 & $1.53$ & 2.75 & 2.46 & $2.61$ & \underline{4.28} & 2.67 & $3.48$ \\
DeepSeek-Template & 2.38 & 2.91 & 2.80 & $2.70$ & 2.75 & 3.68 & \underline{3.84} & $\,\underline{3.42}$ & \underline{1.45} & 1.60 & $1.53$ & 2.74 & 2.46 & $2.60$ & 4.26 & 2.67 & $3.47$ \\
Qwen-Template & \underline{3.01} & \underline{3.21} & \underline{2.87} & $\textbf{3.03}$ & \underline{2.88} & 3.90 & 3.80 & $\,\underline{3.53}$ & \textbf{1.47} & 1.58 & $1.53$ & 2.77 & \underline{2.93} & $\,\underline{2.85}$ & \textbf{4.31} & \underline{3.22} & $3.77$ \\
\midrule
\multicolumn{18}{l}{\textit{Automated generation methods}}  \\
arXiv (HTML) & \textbf{3.70} & 2.23 & 1.34 & $2.42$ & 2.49 & 3.81 & 3.75 & $3.35$ & 1.05 & 1.51 & $1.28$ & 2.72 & 2.65 & $2.69$ & 4.01 & 3.06 & $3.54$ \\
alphaxXiv & \underline{3.43} & 3.01 & 2.91 & $\,\underline{3.12}$ & 2.88 & \underline{3.95} & 3.85 & \textbf{3.56} & 1.25 & 1.61 & $1.43$ & 2.73 & 2.80 & $2.77$ & 4.20 & 3.46 & $\,\underline{3.83}$ \\
\midrule
\rowcolor{blue!8!white} \WebsiteAgent (Our) & 3.06 & \textbf{3.30} & \textbf{2.94} & $\,\underline{3.10}$ & \textbf{2.91} & \textbf{4.02} & \textbf{3.86} &  $\textbf{3.56}$ & 1.39 & \textbf{3.16} &  $\textbf{2.28}$ & \textbf{2.82} & \textbf{3.35} & \textbf{3.09} & \textbf{4.31} & \textbf{3.56} &  $\textbf{3.93}$ \\
\bottomrule
\end{tabular}
\end{adjustbox}
\end{table*}

\subsection{Main Results}

\header{Completeness \& Connectivity.} 
As shown in the left half of Table \redref{tab1}, we evaluate website completeness and connectivity. arXiv-HTML attains high rule-based connectivity but receives 64\% lower human ratings, as it indiscriminately converts every citation into links, inflating metric scores while degrading user experience. alphaXiv shows balanced connectivity by selectively surfacing important links. For completeness, arXiv-HTML preserves verbose text with few images, scoring well with LLM and human judges but poorly on rule-based metrics. In contrast, our \WebsiteAgent achieves 2\% higher LLM-judged completeness than ground truth, demonstrating superior content condensation and balanced layout of text, images, and links. These findings reveal that code-based metrics miss real user experience, motivating our user-centered evaluation next.

\header{Holistic Evaluation.}
As shown in the right half of Table \redref{tab1}, our \WebsiteAgent achieves highest scores across all dimensions. While alphaXiv performs well in completeness and connectivity, it lacks interactive components, scoring 37\% lower than our method in interactivity. Template-based methods effectively guide layout but constrain interactive element generation. Overall, \WebsiteAgent outperforms all generation methods, achieving 91\% of ground truth quality in aesthetics and 94\% in informativeness, with a 59\% improvement in interactivity over alphaXiv.

\begin{table*}[!t]
\caption{PaperQuiz evaluation on the \papertoweb, based on open and closed-source MLLMs. The evaluation metrics include Raw Score and Score with Penalty under two settings: \textit{``Verbatim''} and \textit{``Interpretive''}.}
\label{tab3}
\setlength{\tabcolsep}{2pt}
\centering
\begin{adjustbox}{width=\textwidth,center}
\begin{tabular}{l c c c | c c c | c | c c c c}
\toprule
\textbf{Methods} & \multicolumn{3}{c|}{\textbf{Verbatim}} & \multicolumn{3}{c|}{\textbf{Interpretive}} & \textbf{Avg} & \multicolumn{4}{c}{\textbf{Score with Penalty}} \\
\cmidrule(lr){2-4} \cmidrule(lr){5-7} \cmidrule(lr){9-12}
& open-source $\uparrow$ & closed-source $\uparrow$ & V-Avg  $\uparrow$
& open-source $\uparrow$ & closed-source $\uparrow$ & I-Avg $\uparrow$
& Avg $\uparrow$ & Penalty $\downarrow$ & V\_avg $\uparrow$ & I\_avg $\uparrow$ & Avg $\uparrow$\\
\midrule
\rowcolor{gray!20}
Original Website & 2.94 & 2.14 & 2.54 & 3.81 & 3.09 & 3.45 & 3.00 & 1.43 & 1.11 & 2.02 & 1.57 \\
\midrule
\multicolumn{12}{l}{\textit{Model end-to-end methods}}  \\
GPT-4o & 2.53 & 1.46 & 1.99 & 3.38 & 2.32 & 2.85 & 2.42 & 3.03 & -0.93 & -0.18 & -0.56 \\
Gemini-2.5-flash & 2.60 & 1.59 & 2.10 & 3.14 & 2.72 & 2.93 & 2.52 & 2.18 & -0.19 & 0.71 & 0.24 \\
DeepSeek-V3.2-Exp & 2.55 & 1.54 & 2.00 & 3.21 & 2.55 & 2.88 & 2.44 & 2.26 & -0.26 & 0.62 & 0.18 \\
Qwen3-Coder-480B-A35B & 2.65 & 1.64 & 2.15 & 3.22 & 3.02 & 3.12 & 2.64 & 2.12 & 0.03 & 1.00 & 0.52 \\
\midrule
\multicolumn{12}{l}{\textit{Model end-to-end methods + Template}}  \\
GPT-4o-Template & 2.58 & 1.42 & 2.00 & 3.48 & 2.25 & 2.87 & 2.43 & 2.50 & -0.50 & 0.37 & -0.07 \\
Gemini-Template & 3.62 & 3.36 & 3.49 & 4.40 & \underline{4.45} & 4.42 & 3.96 & 2.01 & 1.48 & 2.41 & 1.95 \\
DeepSeek-Template & 3.55 & 3.19 & 3.37 & 4.11 & 4.25 & 4.18 & 3.78 & \textbf{1.96} & 1.41 & 2.22 & 1.82 \\
Qwen-Template & \underline{3.70} & \underline{3.44} & 3.57 & 4.52 & 4.41 & 4.47 & 4.02 & 2.00 & 1.57 & 2.47 & 2.02 \\
\midrule
\multicolumn{12}{l}{\textit{Automated generation methods}}  \\
arXiv (HTML) & 3.62 & 3.42 & 3.52 & 4.52 & 4.43 & 4.47 & 4.00 & 2.87 & 0.65 & 1.60 & 1.13 \\
alphaxXiv & 3.57 & \textbf{3.60} & \underline{3.58} & \textbf{4.58} & \textbf{4.54} & \textbf{4.56} & \textbf{4.07} & \underline{1.97} & \textbf{1.61} & \textbf{2.59} & \textbf{2.10} \\
\midrule
\rowcolor{blue!8!white} \WebsiteAgent (Our) & \textbf{3.76} & 3.42 & \textbf{3.59} & \underline{4.56} & 4.40 & \underline{4.48} & \underline{4.04} & 2.00 & \underline{1.59} & \underline{2.48} & \underline{2.03} \\
\bottomrule
\end{tabular}
\end{adjustbox}
\end{table*}

\header{PaperQuiz.} As shown in Table~\redref{tab3}, we observe: \textbf{(1)} Without the conciseness penalty, arXiv-HTML scores strongly; once applied, both arXiv-HTML and end-to-end GPT-4o receive large deductions, highlighting the value of concise, engineered sites and supporting website generation as effective context compression. \textbf{(2)} Gemini and Qwen are strong and generally outperform GPT-4o and DeepSeek; templates lift all models—DeepSeek-Template nears Gemini-Template, and Qwen-Template approaches the ground-truth site. \textbf{(3)} Across methods, open-source reader models consistently beat closed-source ones, indicating some open-source MLLMs (e.g., Qwen) can match or exceed closed models on certain visual tasks. \textbf{(4)} \WebsiteAgent achieves best or near-best results across tasks and models, with total information coverage rivaling arXiv-HTML; after the penalty, it still attains the highest overall score. \textbf{(5)} \WebsiteAgent’s penalty remains nontrivial, and the ground-truth site scores lower than expected, likely because it includes many videos and animations; in practice, authors can start from \WebsiteAgent and add multimedia to reach the most desirable design.

\subsection{In-depth Analysis}
\header{Efficiency Analysis.} Figure~\redref{fig:pareto} presents the average token cost per website. Our \WebsiteAgent is highly token-efficient, requiring only \$0.025 to produce a high-quality academic page. By contrast, end-to-end methods are costlier: GPT-4o is about \$0.141 and Gemini about \$0.054 per website. This yields 82\% and 54\% cost reductions, respectively, while maintaining strong page quality and usability. Even template-aided open models around \$0.069 remain 2.8$\times$ more expensive, yet offer no clear advantage. Overall, \WebsiteAgent delivers \emph{state-of-the-art} cost efficiency with high presentation quality.

\begin{figure*}[!t]
  \includegraphics[width=\textwidth]{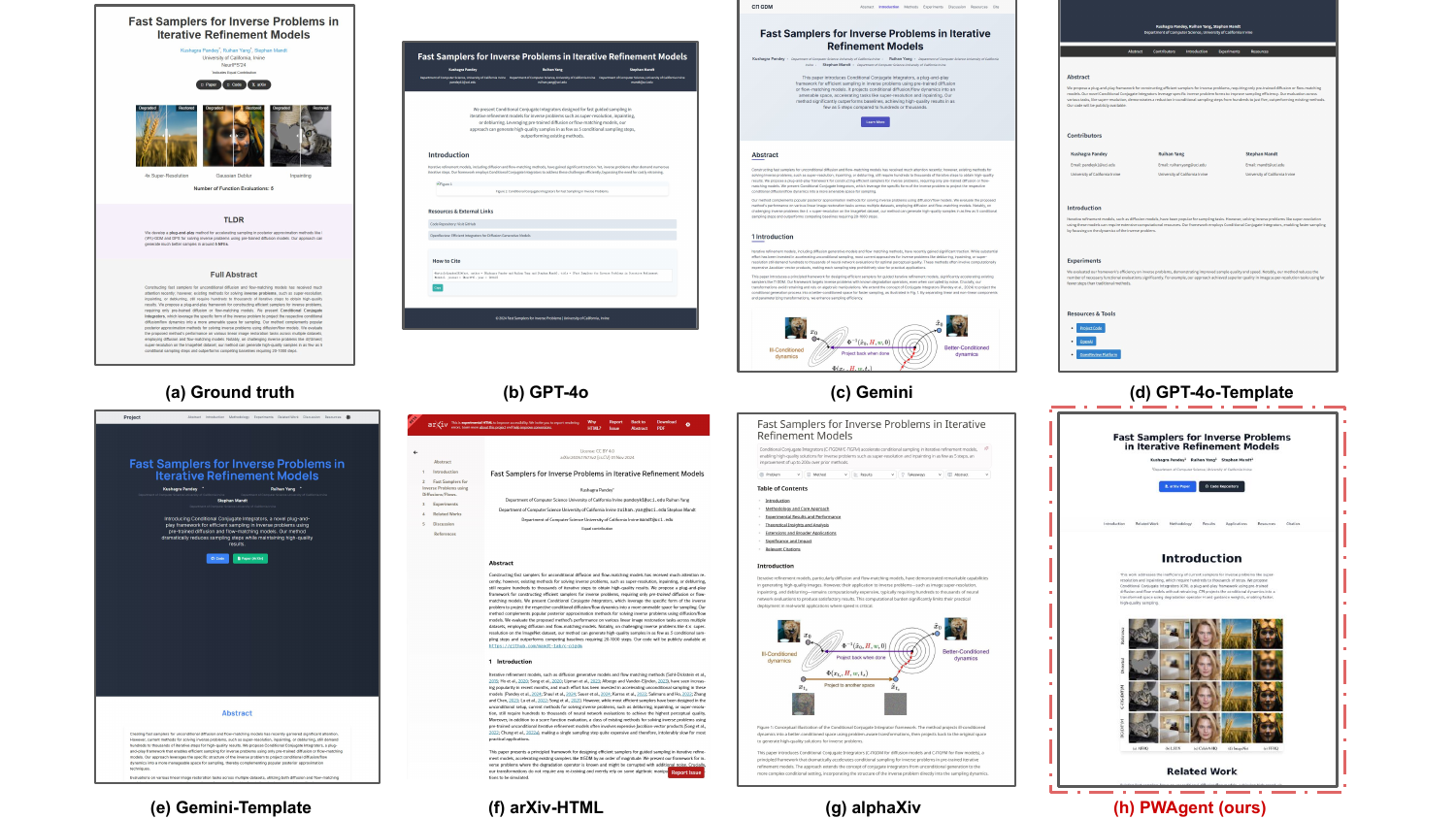}
\caption{Illustration of website variants for the paper generated by different methods. 
GPT-4o fails to cover all components of a paper and amounts to only a simple paradigm; even with template, the sections remain incomplet. The arXiv-HTML is content-rich but is essentially a direct transfer of the original. The alphaXiv method is complete and concise in content, but it lacks a layout paradigm and visual aesthetic quality. Our \agent show interactive and rich multimedia content to enrich presentation quality.
}
  \Description{Case study}
  \label{6}
  \vspace{-1em}
\end{figure*}

\begin{figure}[!t]
  \centering
  \includegraphics[width=\textwidth]{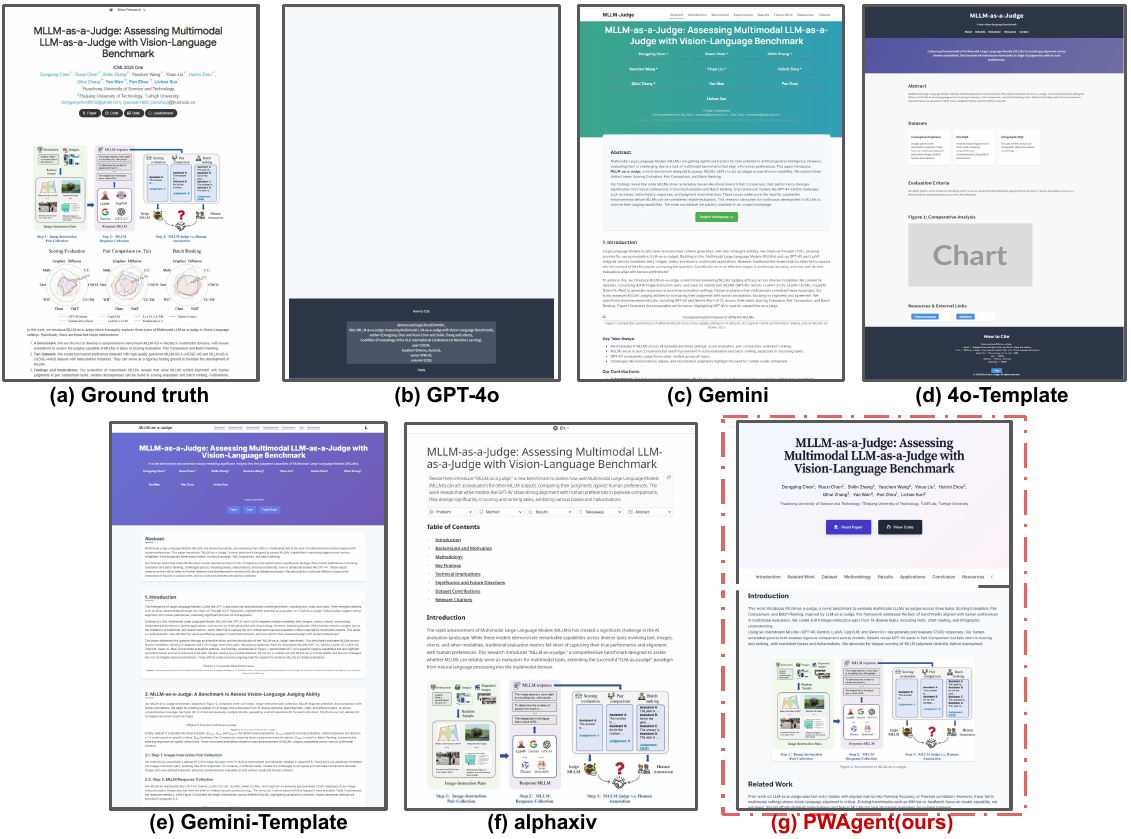}
  \caption{Illustration of website variants for the paper
  ``MLLM-as-a-Judge: Assessing Multimodal LLM-as-a-Judge with Vision-Language Benchmark''\protect\footnotemark generated by different methods.}
  \Description{add_case1}
  \label{fig:add_case1}
  \vspace{-1em}
\end{figure}

\header{Case Study.}
In Figure~\redref{6} and \redref{fig:add_case1}, we present a qualitative comparison of different website baselines for a paper. GPT-4o evidently struggles to generate a structurally coherent HTML webpage from the source PDF, and its content completeness remains poor even when provided with a template. In contrast, the website generated by Gemini appears content-rich at first glance, and its internal structure is significantly improved with a template. However, it suffers from an unbalanced image-to-text ratio with very few visuals, which hinders the reader's ability to systematically understand the project.  The official arXiv-HTML page, while comprehensive, is overly verbose. Although the alphaXiv website is well-illustrated with both images and text, its design is monotonous and lacks aesthetic appeal.
In contrast, our \WebsiteAgent not only preserves the structural integrity of the original paper but also achieves a well-balanced image-to-text ratio. Furthermore, it offers versatile styling and superior aesthetic quality. However, there is still room for improvement when compared to the human-designed version.

\section{Related Work}
\header{HTML Code Generation.}
The field of automated front-end development has seen significant progress, with a primary focus on generating HTML from diverse inputs like screenshots, design prototypes, and natural language descriptions. This research has led to the establishment of several key benchmarks, including Design2Code~\citep{si2024design2code,yang2025multi}, Websight~\citep{laurenccon2024unlocking} and WebCode2M~\citep{gui2025webcode2m}. A variety of code generation strategies have been explored, ranging from direct translation to more structured approaches, such as the divide-and-conquer strategy of DCGen~\citep{wan2024automatically} and the hierarchical generation process used by UICopilot~\citep{gui2025uicopilot}. These technologies have been applied to create mobile UIs\citep{xiao2024interaction2code,zhou2024bridging}, multi-page websites~\citep{wan2024mrweb}, and enhance web design~\citep{xiao2024prototype2code,li2024sketch2code,zhang2024nldesign}, with performance often improved through model fine-tuning~\citep{liang2024waffle}. More recently, multi-agent systems are being increasingly adopted for complex development tasks~\citep{han2024llm,liu2024roleagent}. For example, agentic workflows are now used to convert designs into functional code~\citep{islam2024mapcoder,ding2025frontend}, and some systems assign distinct agents to specific sub-tasks, refining their output through iterative human feedback~\citep{wang2024multimodal}.

\footnotetext{\url{https://mllm-judge.github.io/}}

\header{Automated Processing of Scholarly Articles.}
Early methods for generating derivative content from academic papers primarily relied on template-based~\citep{xu2021neural,qiang2019learning,cheng2024sviptr} or rule-driven models~\citep{huang2022layoutlmv3,lin2023layoutprompter}.Recently, with the maturation of AI agent technology, a substantial body of work has emerged for academic poster generation. A series of methods and benchmarks, including P2P~\citep{sun2025p2p}, Paper-to-Poster~\citep{pang2025paper2poster}, PosterGen~\citep{zhang2025postergen}, CreatiDesign~\citep{zhang2025creatidesign}, PosterCraft~\citep{chen2025postercraft}, and DreamPoster~\citep{hu2025dreamposter}, have explored pipelines for automatically converting papers into posters. These studies demonstrate that through well-designed multi-agent collaboration, the generated posters can achieve high fidelity with human-designed counterparts in terms of layout, content summarization, and visual aesthetics. Similarly, notable progress has been made in presentation slide generation. PresentAgent~\citep{shi2025presentagent}, Preacher~\citep{liu2025preacher}, SciGA~\citep{kawada2025sciga}, and SlideCode~\citep{tang2025slidecoder} introduce specialized datasets, benchmarks, and methodologies. The trend in these task-specific applications is gradually evolving towards broader automated visual design, as exemplified by systems like BannerAgency~\citep{wang2025banneragency} for banner creation and VideoAgent~\citep{wang2024videoagent,fan2024videoagent,soni2024videoagent} for video production.
With the advent of the MCP, researchers have begun to utilize MCP to empower agents for more sophisticated tasks. A prominent example is Paper2Agent~\citep{miao2025paper2agent}, which underscores the potent capabilities of advanced agent systems in handling complex, unstructured academic information.

\section{Conclusion and Discussion}

We introduce \papertoweb, a novel task and benchmark for generating project homepages from academic papers, and identify key challenges faced by current generative models and automated methods in handling long-context and layout-sensitive tasks. Our framework, 
\WebsiteAgent narrows the gap between machine- and human-designed webpages and sets a new efficiency standard for web-based scholarly communication, offering a practical and scalable solution.

While our work represents an initial step toward transforming static papers into exploratory web pages, it primarily aims to define the scope and standards of this emerging area rather than offer a definitive solution. We also propose simple yet multi-dimensional evaluation criteria that lay the groundwork for richer future assessments. Nonetheless, evaluating how multimedia elements contribute to effective academic communication remains an open challenge, which we plan to address through more robust agentic workflows and comprehensive evaluation methods in future work.
We call for continued research on integrating multi-agent reasoning and multimodal understanding to advance the transformation of scholarly communication beyond static formats.

\section*{Acknowledgment}
We thank Gui Yi and Jianuo Huang from Huazhong University of Science and Technology for their valuable input and feedback.

\bibliography{iclr2025_conference} 
\bibliographystyle{iclr2025_conference} 
\newpage
\appendix
\part*{Appendix}  
\addcontentsline{toc}{part}{Appendix}

\begin{center}
  \textbf{\Large Table of Contents}
\end{center}

\startcontents[appendix]
\printcontents[appendix]{l}{1}{\setcounter{tocdepth}{2}}
\newpage

\section{Detailed of rule-base metric}

\subsection{Rule-based Metric for Connectivity}

Connectivity in a web-based academic project can be divided into external links and internal navigations. To quantify this aspect, we first parse the HTML structure of the generated webpage to identify relevant syntactic patterns. Specifically, external links are represented by \texttt{<a href="...">} elements pointing to URLs outside the current domain, while internal navigations are defined by anchor links of the form \texttt{href="\#section-id"}, which reference local sections within the same document.

We record the number of detected external and internal links as $S_{\text{external}}$ and $S_{\text{internal}}$, respectively. For external links, we further employ a URL parser to verify the \textit{validity}, \textit{relevance}, and \textit{accessibility} of each link. Only those URLs that are reachable and contextually relevant to the webpage content are counted toward $S_{\text{external}}$.

The overall rule-based connectivity score $S_{Con}
$ is defined as:
\begin{equation}
S = \frac{S_{\text{external}} + S_{\text{internal}}}{2}
\label{eq:connectivity_score}
\end{equation}

\subsection{Rule-based Metric for Completeness}

\header{Image--Text Balance Prior.}
The Image--Text Balance Prior encodes a heuristic rule: an effective academic project webpage should maintain an approximate balance between visual and textual content, avoiding extremes such as image-only pages or text-dense ``wall-of-text'' layouts. Concretely, we compute the \emph{image--text ratio} of a generated webpage as follows:
\begin{enumerate}
  \item When rendering the full page in a standard viewport, we first measure the area of all containers on the page and calculate the proportion of each container's area occupied by image elements. The image areas are weighted according to the container size.
  \item Text content is treated as the remaining area within each container (excluding images) and is weighted in the same manner by container area proportion.
\end{enumerate}
Finally, the weighted image--text ratios of all containers are aggregated according to the relative area of each container within the entire page, yielding the overall page-level image--text ratio. This approach ensures that a few large images (e.g., full-width banners) and many small icons are appropriately distinguished based on their actual proportions, while the text proportion remains consistent with both container and overall page layout.

\header{Information Efficiency Prior.}
The Information Efficiency Prior rewards concise and information-dense presentations by comparing the generated text length $L$ with the median human-authored length $W$ for comparable sections. In the main text, we introduced the ratio $r = L / W$ together with a scaling factor $\beta$. The median is chosen because human-designed webpages often favor short text supplemented by multimedia, leading to large standard deviations in length; the median better reflects typical requirements while mitigating the influence of extreme cases. The hyperparameter $\beta$ controls the decay rate of the efficiency reward when $L > W$: smaller $\beta$ values impose a stricter penalty on overly verbose text. 
The overall rule-based connectivity score $S_{Con}
$ is defined as:

\begin{equation}
S = \frac{S_{\text{img-txt}} + p(r)}{2}
\label{eq:connectivity_score}
\end{equation}

\section{Human Annotation and Verification Details}
\label{Appendix:human}
The annotation is conducted by 6 authors of this paper independently. The diversity of annotators plays a crucial role in reducing bias and enhancing the reliability of the benchmark. These annotators have knowledge in this domain, with different genders, ages, and educational backgrounds. To ensure the annotators can proficiently mark the data, we provide them with detailed tutorials, teaching them how to evaluate model responses more objectively. Specifically, they are required to give judgments without considering answer lengths, and certain names or positions of the response. Furthermore, we implement cross-validation between different annotators and conduct continuous monitoring to ensure they are maintaining objectivity and fairness. We rate paper webpages with 1--5 integer scores on five indicators: Interactivity, Aesthetic, Informative, Completeness, and Connectivity. Raters inspect the same rendering variant in the tool and score each indicator independently.

\begin{figure}[!t] 
  \centering
  \includegraphics[width=0.8\columnwidth]{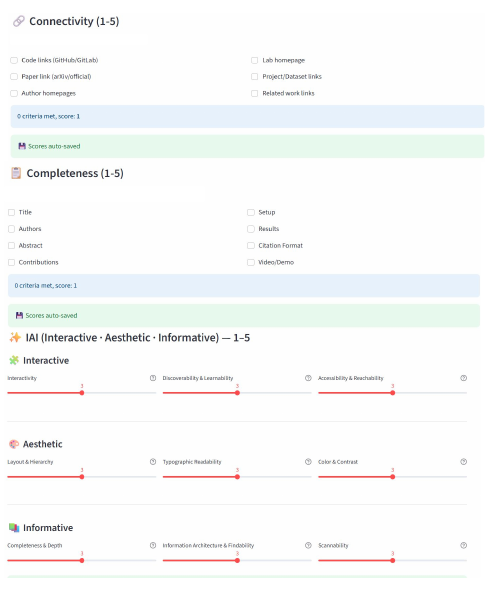}
  \caption{Human annotation instruction}
  \Description{Human annotation instruction.}
  \label{Human annotation instruction}
  \vspace{-1em}
\end{figure}

\subsection{Interactivity}
This metric evaluates the quality and responsiveness of interactions. It also considers discoverability and learnability, where key actions should be obvious and controls self-explanatory. Furthermore, it assesses accessibility and reachability, including keyboard navigation, screen-reader cues, and responsive/mobile usability.

To systematically assess this, we evaluate interactivity across four key areas:

\begin{itemize}[leftmargin=*]

\item \textbf{Basic Interactions:} This criterion covers fundamental dynamic elements that enhance readability. Raters check for common interactions like over effects, expand/collapse sections for content, and clickable tabs.

\item \textbf{Interactive Visualizations:} This assesses dynamic presentations of results. Raters look for interactive charts, comparison sliders, or other elements that allow users to actively explore model outputs.

\item \textbf{Live Demo:} This evaluates the presence of hands-on experiences. Raters check for an embedded online demo or a video that allows users to observe the model's performance directly.

\item \textbf{Navigation Aids:} This focuses on features that improve browsing on long pages. Raters look for tools such as a floating table of contents, quick jump-to-section links, or a back-to-top button.

\end{itemize}

\subsection{Aesthetic}
This dimension focuses on a clear layout and visual hierarchy that guide the user's attention. The WebQuality~\citep{zhang2025webquality} benchmark emphasizes that a well-structured design, which avoids hindering a user's information acquisition, is a cornerstone of quality assessment. This evaluation includes typographic readability, ensured by appropriate font sizes, line height, and stable styling. Color and contrast are also evaluated for being harmonious, accessible, and providing sufficient distinction for all text and interface elements.

We evaluate the aesthetic quality based on four criteria adapted from Paper2Poster~\citep{pang2025paper2poster}:

\begin{itemize}[leftmargin=*] 

\item \textbf{Element Quality:} This criterion assesses the individual visual components on the page. Raters evaluate the clarity and resolution of images, the design quality of illustrations or figures, and whether charts and tables are not only easy to understand but also thoughtfully designed.

\item \textbf{Layout Balance:} This criterion focuses on the overall spatial organization and structure. Raters check for consistent alignment of all elements, reasonable sizing of images, and appropriate spacing between different sections to ensure a clean and flexible layout.

\item \textbf{Engagement and Style:} This criterion evaluates the overall artistic and sensory appeal. Raters assess the consistency and harmony of the color scheme, the readability and appropriateness of the typography, and the creativity of the overall style and its effectiveness in engaging the user.

\item \textbf{Clarity:} Inspired by MLLM as a UI judge~\citep{luera2025mllm}, this criterion evaluates how clear and uncluttered the interface appears, encompassing both textual legibility and overall visual design. A clear interface avoids overwhelming the user with too many UI elements. This extends to the fundamental readability of the text, which should feature a clear, discernible font and sufficient contrast between the text and its background to ensure a comfortable reading experience without strain.

\end{itemize}

\subsection{Informative}
This indicator measures the completeness and depth. It also assesses the information architecture and findability, which are supported by a logical structure, clear labels, and cross-links or search functionality. Scannability, achieved through effective use of headings, bullet points, callouts, and summaries, is another key aspect.

To formalize this assessment, we evaluate the informative quality based on the following three dimensions:

\begin{itemize}[leftmargin=*]

\item \textbf{Logical Flow and Coherence:} Drawing from the criterion in Paper2Poster~\citep{pang2025paper2poster}, this dimension evaluates the overall structure and narrative of the webpage. A high-quality page should present information in a logical sequence that mirrors the research process. The content should be coherent, guiding the reader through the project's story without confusion.

\item \textbf{Depth of Content:} This dimension assesses whether the core sections are explained with sufficient detail and insight. The webpage should provide a substantive discussion of key concepts, methodologies, and findings, demonstrating a thorough understanding and presentation of the research.

\item \textbf{Scannability and Readability:} Raters assess whether the page structures content with visually distinct section titles, applies text formatting like bolding to emphasize key terms, and organizes parallel items or sequential steps into clear lists.

\end{itemize}

\subsection{Completeness}

This metric assesses whether the essential elements of a research webpage are present and sufficiently developed. Recent benchmarks like WebGen-Bench~\citep{lu2025webgen} have highlighted the importance of moving beyond static content to evaluate the generation of truly functional websites.Accordingly, our definition of completeness encompasses not only the presence of core content but also its operational integrity. Raters consider the coverage of core content, the adequacy and coherence of accompanying text and media, and whether the information feels up-to-date and self-contained.

Raters evaluate the presence and thoroughness of items within each dimension.

\begin{itemize}[leftmargin=*]
\item \textbf{Element Completeness:} This dimension evaluates whether the webpage effectively summarizes the fundamental components of the research paper. Drawing from principles in the WebQuality~\citep{zhang2025webquality} dataset, raters assess the presence of essential elements, such as the foundational metadata, a summary of core contributions, descriptions of the experimental setup, and a presentation of key results.

\item \textbf{Rich Media and Artifacts:} This dimension assesses the inclusion of supplementary materials that go beyond static text to enhance understanding and demonstrate practical applications. This includes elements like an embedded video presentation or demo and any interactive visualizations that allow users to explore the data or results.

\item \textbf{Scholarly Utility:} This dimension evaluates features that provide direct practical value to other researchers and facilitate the work's dissemination. This primarily involves tools like an easy-to-copy citation format and clearly labeled links to official resources such as the paper's PDF, source code, or datasets.

\end{itemize}
    
\subsection{Connectivity}
This metric evaluates the richness, relevance, and reliability of outward links. High-quality pages feature working links to code and reproducible artifacts, official paper pages, author or lab websites, datasets, and pertinent related work. Links should be contextually introduced with clear anchor text, free of dead or circular references, and should help readers navigate to deeper resources without friction.

Building upon the work on MRWeb~\citep{wan2024mrweb}, we evaluate connectivity across three dimensions:

\begin{itemize}[leftmargin=*]
\item \textbf{Resource Connectivity:} This dimension assesses the linkage to core research assets that enable reproducibility and deeper engagement. Raters check for direct, functional links to the source code repository, the official paper, and any associated project or data resources.

\item \textbf{Scholarly Context Connectivity:} This dimension measures how well the webpage connects the research to the wider academic landscape. This is primarily evaluated by the presence and quality of links to related work.

\item \textbf{Internal Navigation and Linking:}
This dimension evaluates how effectively the webpage facilitates smooth and intuitive movement between its internal sections. Raters assess the presence and clarity of navigational elements—such as anchored headings, menus, or in-page links—that allow users to easily access key content areas without losing contextual flow.

\end{itemize}

\section{PaperQuiz}

\label{sec:case_study_paperquiz} 

\subsection{QA Dataset Curation.} Each paper PDF is converted to markdown via our PDF parser. We then prompt o3 to generate 50 multiple-choice questions per paper, where we have 25 verbatim and 25 interpretive questions as follows:

\begin{itemize}[leftmargin=*]
    \item \textbf{Verbatim questions (25):} directly answerable from the paper text, covering 13 orthogonal content aspects (e.g., objectives, methodology, key results).
    \item \textbf{Interpretive questions (25):} requires high-level comprehension beyond verbatim text, spanning 10 conceptual dimensions (e.g., motivation, contribution synthesis, implication analysis).
\end{itemize}

The following is a prompt example of 25 Verbatim questions and 25 Interpretive questions, generated by GPT-o3.

\begin{tcolorbox}[
  enhanced,
  breakable,
  sharp corners=south,
  arc=4mm,
  boxrule=0.8pt,
  colframe=black!60,
  colback=gray!5,
  width=\linewidth,
  left=8pt,right=8pt,top=8pt,bottom=6pt,
  before skip=8pt, after skip=8pt,
  title=Prompt: Generated Verbatim Questions,
  colbacktitle=gray!20,
  coltitle=black,
  fonttitle=\bfseries,
  attach boxed title to top left={yshift=-3pt,xshift=6pt},
  boxed title style={
    enhanced,
    boxrule=0pt,
    left=6pt, right=6pt, top=2pt, bottom=2pt,
    arc=3mm
  }
]
{\raggedright 
\textbf{System\_prompt:}\par
 You are a highly precise Question-Generation agent for academic project websites.
  Your task is to read the supplied Markdown text
  and produce a structured set of exactly 25 multiple\-choice QA items. 
  Your primary goal is to strictly adhere to a mandatory Question Distribution Plan and a set of critical formatting rules. 
  Failure to follow these rules precisely will result in an invalid output. 
  The answers to your questions must be located verbatim or almost verbatim in the provided text. 
  The questions must be suitable for website visitors: avoid deep theoretical proofs, reference lists, or citation minutiae.\par

\medskip
\textbf{Instructions:}\par
You MUST generate questions according to the following to ensure all aspects are covered and the total is exactly 25 questions.

\begin{itemize}
  \item \textbf{Locate Fact:} Find a specific, clear factual statement, number, or detail in the `document\_markdown.
  \item \textbf{Classify Aspect:} Critically determine which single aspect (from A-M) this fact *most accurately* represents. Be extremely strict in your classification.
  \item \textbf{Formulate Question:} Based ONLY on the located fact, create a clear, answerable-from-text question.
  \item \textbf{Create Options:} Write the correct answer and three high-quality distractors as defined in the rules below.
\end{itemize}

\textbf{Aspect Definitions \& Special Instructions:}

You will generate questions for the following aspects:

\begin{itemize}
  \item A. Research domain \& background context.
  \item B. Central problem / motivation / research gap
  \item C. Primary goal, hypothesis, or research question
  \item D. Key contributions or novelty statements
  \item E. Overall methodology or workflow (summarized)
  \item F. Qualitative insights or illustrative examples \par
  .....
\end{itemize}

\textbf{MANDATORY Formatting Rules:}

Each question object MUST have exactly four options, labelled \texttt{"A."}, \texttt{"B."}, \texttt{"C."}, and \texttt{"D."}. Do not generate more or fewer than four. The "aspect" key is required and must contain a single letter from the list above.

\noindent\textbf{Final Pre-Output Check:} Before providing the final JSON, mentally perform this check:
\begin{itemize}
  \item Is the total number of questions EXACTLY 25?
  \item Is the Question Distribution Plan followed perfectly?
  \item Does EVERY single question have EXACTLY four options?
  \item Is every question accurately classified with its \texttt{aspect} and do they follow all special instructions?
  \item If any check fails, you must restart and correct the errors.
\end{itemize}

\textbf{document\_markdown}: \\
  \{\{ document\_markdown \}\}

}

\end{tcolorbox}
\begin{tcolorbox}[
  enhanced,
  breakable,
  sharp corners=south,
  arc=4mm,
  boxrule=0.8pt,
  colframe=black!60,
  colback=gray!5,
  width=\linewidth,
  left=8pt,right=8pt,top=8pt,bottom=6pt,
  before skip=8pt, after skip=8pt,
  title=Prompt: Generated Interpretive Questions,
  colbacktitle=gray!20,
  coltitle=black,
  fonttitle=\bfseries,
  attach boxed title to top left={yshift=-3pt,xshift=6pt},
  boxed title style={
    enhanced,
    boxrule=0pt,
    left=6pt, right=6pt, top=2pt, bottom=2pt,
    arc=3mm
  }
]
{\raggedright 
\textbf{System\_prompt:}\par
You are a highly precise Question\-Generation agent for academic project websites.
Your task is to read the supplied Markdown text and produce a structured set of exactly 25 multiple-choice QA items. Your primary goal is to strictly adhere to a mandatory Question Distribution Plan and a set of critical formatting rules. The answers to your questions must be located verbatim or almost verbatim in the provided text. The questions must be suitable for website visitors.\par

\medskip
\textbf{Instructions:}\par
For each question you generate, you MUST follow these mental steps:

\begin{itemize}
  \item \textbf{Locate Fact:} Find a specific, clear factual statement, number, or detail in the document\_markdown.
  \item \textbf{Classify Aspect:} Critically determine which single aspect (from A--M) this fact most accurately represents. Be extremely strict in your classification.
  \item \textbf{Formulate Question:} Based ONLY on the located fact, create a clear, answerable-from-text question.
  \item \textbf{Create Options:} Write the correct answer and three high-quality distractors as defined in the rules below.
\end{itemize}

\textbf{Aspect Definitions \& Special Instructions:}

You will generate questions for the following aspects:

\begin{itemize}
  \item A. Title \& authorship (title, author names, affiliations, keywords): For questions about author names, the incorrect options (distractors) MUST be fabricated but plausible-sounding names. Do not use real names from other contexts.
  \item B. Motivation / problem statement / research gap
  \item C. Objectives or hypotheses
  \item D. Dataset(s) or experimental materials
  \item E. Methodology (algorithms, model architecture, workflow steps)
  \item F. Key parameters or hyper-parameters (values, settings) \par
  .....
\end{itemize}

\textbf{MANDATORY Formatting Rules:}

Each question object MUST have exactly four options, labelled \texttt{"A."}, \texttt{"B."}, \texttt{"C."}, and \texttt{"D."}. Do not generate more or fewer than four. The "aspect" key is required and must contain a single letter from the list above.

\noindent\textbf{Final Pre-Output Check:} Before providing the final JSON, mentally perform this check:
\begin{itemize}
  \item Is the total number of questions EXACTLY 25?
  \item Is the Question Distribution Plan followed perfectly?
  \item Does EVERY single question have EXACTLY four options?
  \item Is every question accurately classified with its \texttt{aspect} and do they follow all special instructions?
  \item If any check fails, you must restart and correct the errors.
\end{itemize}

\noindent\textbf{Adhere to Mandatory JSON Format:}\par

\textbf{document\_markdown}: \\
  \{\{ document\_markdown \}\}

}
\end{tcolorbox}

\subsection{Evaluation Workflow.} For each website snapshot, we query six MLLMs reader models to answer curated questions. These models include three open-source models (LLaVA-OneVision-Qwen2-7B-ov-hf, DeepSeek-V3.2-Exp, and Qwen3-Coder-480B-A35B) and three closed-source models (o1, Gemini 2.5 Flash, and Grok Code Fast 1). Their outputs are evaluated according to two enforced rules:
\begin{itemize}[leftmargin=*]
    \item \textbf{No external knowledge.} Models must base answers solely on information present in the website snapshot.
    
    \item \textbf{Visual citation.} Each answer must include a reference to the website region supporting it (e.g., ``See the `Results' section''); if no region contains the answer, the model responds ``NA.''
\end{itemize}

\begin{tcolorbox}[
  enhanced,
  breakable,
  sharp corners=south,
  arc=4mm,
  boxrule=0.8pt,
  colframe=black!60,
  colback=gray!5,
  width=\linewidth,
  left=8pt,right=8pt,top=8pt,bottom=6pt,
  before skip=8pt, after skip=8pt,
  title=Prompt: Answer Qusetions,
  colbacktitle=gray!20,
  coltitle=black,
  fonttitle=\bfseries,
  attach boxed title to top left={yshift=-3pt,xshift=6pt},
  boxed title style={
    enhanced,
    boxrule=0pt,
    left=6pt, right=6pt, top=2pt, bottom=2pt,
    arc=3mm
  }
]
{\raggedright
\textbf{System\_prompt:}\par
You are an answering agent. You will be provided with: 
\begin{itemize}
  \item An image of a project website snapshot.
  \item A JSON object called ``questions'' which contains multiple questions. Each question has four possible answers: A, B, C, or D.
\end{itemize}

Your goal is to analyze the website snapshot thoroughly and answer each question based on the information it provides. You should \textbf{NOT} use any external knowledge or context beyond the website snapshot image. You must rely solely on the content of the website snapshot to answer the questions.

\medskip
For each question:
\begin{itemize}
  \item If you find enough evidence in the website snapshot to decide on a specific option (A, B, C, or D), then choose that option. Also include a brief reference to the part of the webpage that supports your answer (e.g., ``Top-left text'', ``Header section'', etc.).
  \item If the website snapshot does not offer sufficient information to confidently choose any of the options, respond with ``NA'' for both the answer and the reference.
\end{itemize}

\medskip
Your final output must be returned as a JSON object. For each question, the structure should be:

\begin{verbatim}
"Question N": {
  "answer": "A" | "B" | "C" | "D" | "NA",
  "reference": "<short description or 'NA'>"
}
\end{verbatim}

\medskip
\textbf{Template:}\par
Follow these steps to create your response:
\begin{enumerate}
  \item Study the website snapshot image along with the ``questions'' provided.
  \item For each question:
    \begin{itemize}
      \item Decide if the website snapshot clearly supports one of the four options (A, B, C, or D). If so, pick that answer.
      \item Otherwise, if the website snapshot does not have adequate information, use ``NA'' for the answer.
    \end{itemize}
  \item Provide a brief reference indicating where on the webpage you found the answer. If no reference is available (i.e., your answer is ``NA''), use ``NA'' for the reference too.
  \item Format your output strictly as a JSON object with this pattern:
\begin{verbatim}
{
  "Question 1": {
    "answer": "X",
    "reference": "some reference or 'NA'"
  },
  "Question 2": {
    "answer": "X",
    "reference": "some reference or 'NA'"
  },
  ...
}
\end{verbatim}
  \item Do not include any explanations or extra keys beyond the specified structure.
  \item You must provide an answer entry for all questions in the ``questions'' object.
\end{enumerate}

\medskip
\textbf{Example Output:}

\medskip
\textbf{Questions:}
\begin{verbatim}
{{questions}}
\end{verbatim}
}

\end{tcolorbox}

\subsection{Case Study for PaperQuiz}
Here we provide a simple Q\&A example of PaperQuiz.
\begin{tcolorbox}[
  enhanced,
  breakable,
  sharp corners=south,
  arc=4mm,
  boxrule=0.8pt,
  colframe=black!60,
  colback=gray!5,
  width=\linewidth,
  left=8pt,right=8pt,top=8pt,bottom=6pt,
  before skip=8pt, after skip=8pt,
  title=PaperQuiz Example,
  colbacktitle=gray!20,
  coltitle=black,
  fonttitle=\bfseries,
  attach boxed title to top left={yshift=-3pt,xshift=6pt},
  boxed title style={
    enhanced,
    boxrule=0pt,
    left=6pt, right=6pt, top=2pt, bottom=2pt,
    arc=3mm
  }
]
{\raggedright 
{\small
\begin{Verbatim}[breaklines=true, breaksymbol=]
{
  "questions": {
    "Question 1": {
      "question": "What is the full title of the paper discussed in the document?",
      "options": [
        "A. Universal Audio-Video Diffusion Networks for Multimodal Synthesis",
        "B. Multisensory Diffusion: A Joint Model for Sound and Vision",
        "C. Cross-Modal Transformer: Unified Audio and Video Generation",
        "D. A Versatile Diffusion Transformer with Mixture of Noise Levels for Audiovisual Generation"
      ]
    },
    ...
    "Question 25": {
      "question": "In the context of this paper, what does the term \"time-segment\" specifically refer to?",
      "options": [
        "A. An entire training epoch",
        "B. One complete diffusion timestep in noise addition",
        "C. A full audio clip of any length",
        "D. A single unit in the temporal dimension such as a video frame"
      ]
    }
  },

  "answers": {
    "Question 1": "D. A Versatile Diffusion Transformer with Mixture of Noise Levels for Audiovisual Generation",
    ...
    "Question 25": "D. A single unit in the temporal dimension such as a video frame"
  },

  "aspects": {
    "Question 1": "A",
    ...
    "Question 25": "M"
  },

  "understanding": {
    "questions": {
      "Question 1": {
        "question": "What multimodal generation challenge is identified as still open in the paper's introduction?",
        "options": [
          "A. Inferring audio labels from isolated spectrogram snapshots",
          "B. Classifying large multimodal datasets into predefined categories",
          "C. Producing single high-resolution images from textual captions",
          "D. Generating sequences across multiple modalities such as video and audio"
        ]
      },
      ...
      "Question 25": {
        "question": "Which unified approach is claimed by the authors to enable a single model to generate and manipulate sequences across modalities and time?",
        "options": [
          "A. The mixture of noise levels strategy introduced in this paper",
          "B. An unsupervised text summarization algorithm",
          "C. A rule-based system for audio classification",
          "D. A curriculum learning schedule for GANs"
        ]
      }
    },

    "answers": {
      "Question 1": "D. Generating sequences across multiple modalities such as video and audio",
      ...
      "Question 25": "A. The mixture of noise levels strategy introduced in this paper"
    },

    "aspects": {
      "Question 1": "A",
      ...
      "Question 25": "J"
    }
  }
}
\end{Verbatim}
}

}
\end{tcolorbox}

\section{Prompt Template}
\subsection{Baseline Template}
We exhibit the prompt templates used to generate end-to-end model generation baselines. When incorporating the template from the popular Nerfies academic website~\citep{park2021nerfies}, you only need to include this template as part of the prompt.

\begin{tcolorbox}[
  enhanced,
  breakable,
  sharp corners=south,
  arc=4mm,
  boxrule=0.8pt,
  colframe=black!60,
  colback=gray!5,
  width=\linewidth,
  left=8pt,right=8pt,top=8pt,bottom=6pt,
  before skip=8pt, after skip=8pt,
  title=Prompt: Baseline LLM Generation,
  colbacktitle=gray!20,
  coltitle=black,
  fonttitle=\bfseries,
  attach boxed title to top left={yshift=-3pt,xshift=6pt},
  boxed title style={
    enhanced,
    boxrule=0pt,
    left=6pt, right=6pt, top=2pt, bottom=2pt,
    arc=3mm
  }
]
{\raggedright 
\textbf{System\_prompt:}\par
  You are a document-to-website generation agent and n expert full-stack web developer and UI/UX designer specializing in creating beautiful, modern, and interactive academic project websites. Your task is to generate a complete, production-ready website based on research paper content and visual asset allocations.
  Your task is to read the supplied Markdown text and design a professional, visually appealing academic conference website by generating an HTML file. Follow the guidelines below precisely.\par

\begin{itemize}
  \item Is visually stunning and modern with a professional, clean, and academic design.
  \item Has rich interactivity and smooth animations.
  \item Effectively presents research content in an engaging way.
  \item Integrates external links and resources strategically.
  \item Uses advanced CSS and JavaScript for enhanced user experience.
\end{itemize}

\medskip
\textbf{Instructions:}\par
 You are creating a complete, beautiful, and interactive website for an academic research project. This is NOT a simple static page - it should be a sophisticated, modern web application with rich interactivity.
  Your task is to read the supplied Markdown text and design a professional, visually appealing academic conference website by generating an HTML file.

\begin{itemize}
  \item \textbf{Design Requirements}
  \begin{itemize}
    \item \textbf{Visual Design}
    \begin{itemize}
      \item Modern, professional, academic aesthetic.
      \item Sophisticated color scheme (dark/light themes with multiple color variations).
      \item Professional typography with hierarchy and multiple font weights.
      \item Smooth animations and transitions with multiple animation types.
      \item Interactive elements and hover effects with complex state changes.
      \item Professional spacing and layout with multiple breakpoints.
      \item Advanced visual effects (shadows, gradients, transforms).
      \item \textbf{Background Style:} Avoid background images (especially in hero section); prefer solid colors such as \#2d3748 (dark gray) or \#ffffff (white) or subtle gradients. Do not fetch images from external sources like Unsplash.
    \end{itemize}

    \item \textbf{Layout Structure}
    \begin{itemize}
      \item Hero section with project title, authors, and key highlights.
      \item Multi-level navigation with smooth scrolling and active state indicators.
      \item Content sections with dynamic layouts based on importance.
      \item Interactive visualizations and image galleries with lightbox and carousel.
      \item External resources section with categorized link placement.
      \item Footer with information, social links, and contact details.
      \item Sidebar navigation with quick links and progress indicators.
      \item Multiple columns and grid layouts.
      \item Card-based content presentation.
    \end{itemize}

    \item \textbf{Interactivity Features}
    \begin{itemize}
      \item Smooth scrolling navigation with progress bars and scroll indicators.
      \item Interactive image galleries with lightbox, zoom, and slideshow.
      \item Animated counters and number transitions.
      \item Hover effects and micro-interactions.
      \item Responsive navigation menu with hamburger and dropdowns.
      \item Loading animations and skeleton screens.
      \item Interactive charts and visualizations with tooltips.
      \item Modal dialogs and popup windows.
      \item Form validation and interactive feedback.
      \item Search with autocomplete.
      \item Dark/light theme toggle with transitions.
    \end{itemize}

    \item \textbf{External Links Integration}
    \begin{itemize}
      \item Place important links strategically within content.
      \item Dedicated “Resources \& Tools” section with categories.
      \item Integrate links naturally in context.
      \item Attractive buttons for external links with hover effects.
      \item Provide descriptive context for each resource.
    \end{itemize}
  \end{itemize}

  \item \textbf{Technical Requirements}
  \begin{itemize}
    \item \textbf{CSS Requirements}
    \begin{itemize}
      \item Advanced animations and transitions with varied timing.
      \item Responsive design for mobile, tablet, and desktop.
      \item CSS Grid and Flexbox layouts.
      \item CSS variables for theming.
      \item Advanced selectors and pseudo-elements.
      \item Center single and multiple images responsively (max 3 per row).
      \item Smooth scrolling and scroll animations.
      \item Hover effects and micro-interactions.
      \item Professional color schemes with multiple variations.
      \item Advanced typography with clear hierarchy.
    \end{itemize}

    \item \textbf{JavaScript Requirements}
    \begin{itemize}
      \item Modern ES6+ syntax with error handling.
      \item Interactive image galleries with lightbox.
      \item Smooth scrolling navigation and progress indicators.
      \item Mobile menu with animations.
      \item Intersection Observer for scroll animations.
      \item Local storage for user preferences.
      \item Form validation and interactive feedback.
      \item Performance optimization and error handling.
      \item Advanced image handling and gallery functionality.
    \end{itemize}

    \item \textbf{Critical JavaScript Best Practices (MUST FOLLOW)}
    \begin{itemize}
      \item \textbf{DOM Element Access Timing:} All DOM element access must occur within a \texttt{DOMContentLoaded} listener.
      \item \textbf{Intersection Observer Setup:} 
      \begin{itemize}
        \item Set up observer before adding classes.
        \item Observe elements immediately after adding \texttt{fade-in} class.
        \item Never query \texttt{.fade-in} elements before setup.
        \item Example: \texttt{element.classList.add('fade-in'); observer.observe(element);}
      \end{itemize}
      \item \textbf{Event Listener Safety:} Always verify element existence before adding listeners.
      \item \textbf{Animation Class Management:} Ensure fade-in classes start invisible (\texttt{opacity: 0}) and become visible (\texttt{opacity: 1}) when animated.
      \item \textbf{Function Organization:} Wrap DOM-dependent code in initialization functions triggered by \texttt{DOMContentLoaded}.
    \end{itemize}
  \end{itemize}

  \item \textbf{Final Checklist}
  \begin{itemize}
    \item Header includes title, authors, and affiliations.
    \item Images sized using responsive CSS (\texttt{width: 100\%}).
    \item Dedicated “Resources \& External Links” section with clickable URLs.
    \item Each URL accompanied by description.
    \item Preserve all original text content.
    \item Images fit properly within containers.
    \item Lists rendered as responsive grids.
  \end{itemize}

  \item \textbf{Critical Checks}
  \begin{itemize}
    \item Consistent, professional typography using fonts like Inter or Manrope.
    \item Prominent author display below title with affiliations.
    \item “How to Cite” section with BibTeX and “Copy” button.
    \item No fixed image sizes in HTML; control via CSS (\texttt{w-full}).
    \item Implement Scroll-Spy in navigation.
    \item Encourage interactive demos over static images.
    \item Add elegant hover and scaling effects to all buttons.
  \end{itemize}
\end{itemize}

\textbf{document\_markdown}: \\
  \{\{ document\_markdown \}\}

\textbf{jinja\_args}: \\
  - document\_markdown
}

\end{tcolorbox}

\subsection{Parsing Template}
We present the prompt templates used for paper deconstruction: (1) the prompt for paper summarizeing, and (2) the prompt for image and table filtering.

\begin{tcolorbox}[
  enhanced,
  breakable,
  sharp corners=south,
  arc=4mm,
  boxrule=0.8pt,
  colframe=black!60,
  colback=gray!5,
  width=\linewidth,
  left=8pt,right=8pt,top=8pt,bottom=6pt,
  before skip=8pt, after skip=8pt,
  title=Prompt: Paper Summarizeing,
  colbacktitle=gray!20,
  coltitle=black,
  fonttitle=\bfseries,
  attach boxed title to top left={yshift=-3pt,xshift=6pt},
  boxed title style={
    enhanced,
    boxrule=0pt,
    left=6pt, right=6pt, top=2pt, bottom=2pt,
    arc=3mm
  }
]
{\raggedright
\begin{itemize}
  \item You are the author of the paper, and you will create a comprehensive content summary for a project website. Your task is to extract and expand the key information from the research paper to create detailed, informative content for each section.
  
  \item \textbf{IMPORTANT REQUIREMENTS:}
  \begin{itemize}
    \item \textbf{Dual Constraint Adherence}: Each section must strictly meet BOTH of the following constraints.
    \item \textbf{Content Richness}: On the premise of ensuring the character and sentence counts are not exceeded, each section must be rich with substantial detail.
    \item \textbf{Information Completeness}: Include comprehensive coverage of all paper content, not just summaries.
    \item \textbf{Website Depth}: Provide enough detail for website visitors to fully understand the research without reading the paper.
    \item \textbf{Technical Thoroughness}: Explain technical concepts, methods, and results in detail.
  \end{itemize}

  \item \textbf{CONTENT STRUCTURE FOR EACH SECTION:}
  
  The constraints below apply to every section (Introduction, Related Work, etc.).
  \begin{itemize}
    \item \textbf{Introduction}: Write sentences covering the research background, core motivation, challenges, main contributions, and a general overview.
    \item \textbf{Related Work}: Write sentences covering existing approaches, their detailed limitations, and the specific gaps in current research.
    \item \textbf{Dataset Overview}: Write sentences covering the dataset's composition, key features, core statistics, comparisons with other datasets, and its detailed characteristics.
    \item \textbf{Methodology/Approach}: Writh sentences covering core technical details, key algorithms, the implementation process, and the specific methods used.
    \item \textbf{Results/Evaluation}: Write sentences covering the experimental setup, detailed core results, analysis of the results, and comprehensive performance comparisons.
    \item \textbf{Applications}: Write sentences covering specific use cases, benefits, practical application scenarios, and representative examples.
    \item \textbf{Conclusion}: Write sentences covering a summary of the research, reiterating the contributions, pointing out limitations, and providing a detailed outlook on future work.
  \end{itemize}

  \item \textbf{OUTPUT FORMAT:}
  
  Generate a JSON object with the following structure:

  \item \textbf{CONTENT GUIDELINES:}
  
  On the premise of ensuring the character and sentence counts are not exceeded, please adhere to the following as much as possible:
  \begin{itemize}
    \item \textbf{Expand information}: Provide comprehensive coverage of paper content.
    \item \textbf{Include specific numbers}: Use actual statistics, dimensions, and measurements from the paper.
    \item \textbf{Be thorough and detailed}: Explain concepts, methods, and results in depth.
    \item \textbf{Explain significance}: Why is this important? What problems does it solve?
    \item \textbf{Compare and contrast}: How does this compare to existing approaches?
    \item \textbf{Future implications}: What are the broader impacts and applications?
    \item \textbf{Provide examples}: Include concrete examples and use cases.
    \item \textbf{Maintain technical depth}: Do not oversimplify technical concepts.
  \end{itemize}

  \item Paper content to analyze:
  \begin{verbatim}
  {{ markdown_document }}
  \end{verbatim}
\end{itemize}
}

\end{tcolorbox}

\begin{tcolorbox}[
  enhanced,
  breakable,
  sharp corners=south,
  arc=4mm,
  boxrule=0.8pt,
  colframe=black!60,
  colback=gray!5,
  width=\linewidth,
  left=8pt,right=8pt,top=8pt,bottom=6pt,
  before skip=8pt, after skip=8pt,
  title=Prompt: Image/Table Filtering,
  colbacktitle=gray!20,
  coltitle=black,
  fonttitle=\bfseries,
  attach boxed title to top left={yshift=-3pt,xshift=6pt},
  boxed title style={
    enhanced,
    boxrule=0pt,
    left=6pt, right=6pt, top=2pt, bottom=2pt,
    arc=3mm
  }
]
{\raggedright
\begin{itemize}
  \item You are an assistant that reviews a research paper's content (\texttt{json\_content}), along with corresponding \texttt{image\_information} and \texttt{table\_information}. Your task is to filter out any image or table entries that are irrelevant to the content described in \texttt{json\_content}, specifically for creating a project website.
  
  \item Specifically:
  \begin{itemize}
    \item Read through the full research paper data described in \texttt{json\_content}.
    \item Examine each entry within \texttt{image\_information} and \texttt{table\_information}.
    \item Decide if each entry is relevant for a project website based on its caption, path, or any other information provided.
    \begin{itemize}
      \item For example, if an image has a caption that obviously does not fit into any section or does not relate to the paper's content outline, deem it "unimportant."
      \item Consider which images/tables would be most valuable for a project website.
    \end{itemize}
    \item Keep all images/tables that are relevant to the project website (i.e., related to the topics, sections, or discussions mentioned in \texttt{json\_content}).
    \item Do not impose any artificial quantity limits—include every visual element that enhances understanding of the research.
    \item Produce an output containing just two keys: "image\_information" for the filtered images, and table\_information" for the filtered tables. Each of these keys should map to an array of filtered objects.
  \end{itemize}
  \begin{itemize}
    \item The user will provide JSON: 
      \begin{itemize}
        \item \texttt{"json\_content"}: The content of the research paper (sections, text, etc.)
        \item \texttt{"image\_information"}: A dict of images (each with caption, path, size constraints)
        \item \texttt{"table\_information"}: A dict of tables (each with caption, path, size constraints)
      \end{itemize}

    \item Your task:
    \begin{itemize}
      \item Read the research paper outline (\texttt{json\_content}).
      \item Filter \texttt{image\_information} and \texttt{table\_information} so that only entries relevant to the project website content remain.
      \item Relevance is determined by matching or relating captions to the paper’s sections or content.
      \item Consider which visual elements would be most valuable for a project website (e.g., methodology diagrams, result charts, data summaries).
      \item If an image or table does not clearly match or support any content in \texttt{json\_content}, remove it.
      \item Keep all relevant visual elements—do not limit the quantity artificially.
    \end{itemize}
  \end{itemize}
  \item You must output valid JSON containing only:
{\small
\begin{verbatim}
{
  "image_information": {...},
  "table_information": {...}
}
\end{verbatim}
}

  \item Template Instructions:

  \item Please provide only the JSON object as your final output.

  \begin{verbatim}
  json_content:
  {{ json_content }}

  image_information:
  {{ image_information }}

  table_information:
  {{ table_information }}
  \end{verbatim}

  \item Jinja arguments:
  \begin{verbatim}
  - image_information
  - table_information
  - json_content
  \end{verbatim}
\end{itemize}
}
\end{tcolorbox}

\subsection{Orchestrating Template}

We introduce the prompt templates that guide the Agent-Driven Iterative Refinement procedure.
\begin{tcolorbox}[
  enhanced,
  breakable,
  sharp corners=south,
  arc=4mm,
  boxrule=0.8pt,
  colframe=black!60,
  colback=gray!5,
  width=\linewidth,
  left=8pt,right=8pt,top=8pt,bottom=6pt,
  before skip=8pt, after skip=8pt,
  title=Prompt for MLLM as Orchestrator,
  colbacktitle=gray!20,
  coltitle=black,
  fonttitle=\normalsize,
  attach boxed title to top left={yshift=-3pt,xshift=6pt},
  boxed title style={
    enhanced,
    boxrule=0pt,
    left=6pt, right=6pt, top=2pt, bottom=2pt,
    arc=3mm
  }
]
\raggedright

\textbf{System\_message}:  \\
You are an expert web developer and UI/UX designer with extensive experience in analyzing website layouts, visual design, and user experience. Your task is to analyze website screenshots and provide targeted recommendations for improvement.

Your mission is to first classify the type of web component shown in a screenshot, and then analyze and optimize it based on a deep understanding of modern design systems and principles, using a protocol tailored to that specific component type. You must provide surgically precise feedback to guide code fixes.

\medskip

\textbf{Core Mission}

\begin{itemize}
  \item \textbf{Protocol for "Navigator"}\\
    Focus: Ensure clarity, usability, and responsiveness in navigation elements.\\[4pt]
    1. Component Flow \& Alignment\\
       Diagnosis: Are navigation links properly aligned?\\
       Action: Suggest adjusting flexbox/grid properties (justify-content, gap) or applying uniform margins.\\[4pt]
    2. Typography \& Readability\\
       Diagnosis: Are the link labels easy to read?\\
       Action: Recommend increasing font size, adjusting font weight, or modifying colors for contrast.\\[4pt]
    \dots
  \item \textbf{Protocol for "Header/Hero"}\\
    Focus: Maximize visual impact, establish a clear hierarchy, and communicate the primary purpose.\\[4pt]
    1. Visual Hierarchy \& Flow\\
       Diagnosis: Is the main heading prominent?\\
       Action: Adjust font sizes or positioning to create a clear focal point.\\[4pt]
    2. Image Dominance \& Sizing\\
       Diagnosis: Does the background image enhance or overwhelm content?\\
       Action: Suggest constraining height or applying a semi-transparent overlay.\\[4pt]
    \dots
  \item \textbf{Protocol for "Content Block" \& "Component/Card"}\\
    Focus: Ensure logical structure, effortless readability, and visual consistency.\\[4pt]
    1. Component Flow \& Layout\\
       Diagnosis: Are grouped elements laid out logically?\\
       Action: Suggest using CSS Flexbox or Grid for adaptive alignment.\\[4pt]
    2. Typography \& Readability\\
       Diagnosis: Is the text comfortable to read?\\
       Action: Recommend adjusting line-height and ensuring adequate contrast.\\[4pt]
    \dots
\end{itemize}

\medskip

Response format:  \\
{\small
\begin{verbatim}
{
  "is_needed_to_fix": true/false,
  "category": "The identified category of the component: Navigator | 
  Header/Hero | Content Block | Component/Card",
  "fix_suggest": "Detailed analysis and suggestions"
}
\end{verbatim}
}
\end{tcolorbox}

\section{More Examples of Case Study}

\begin{figure}[H]
  \centering
  \includegraphics[width=\textwidth]{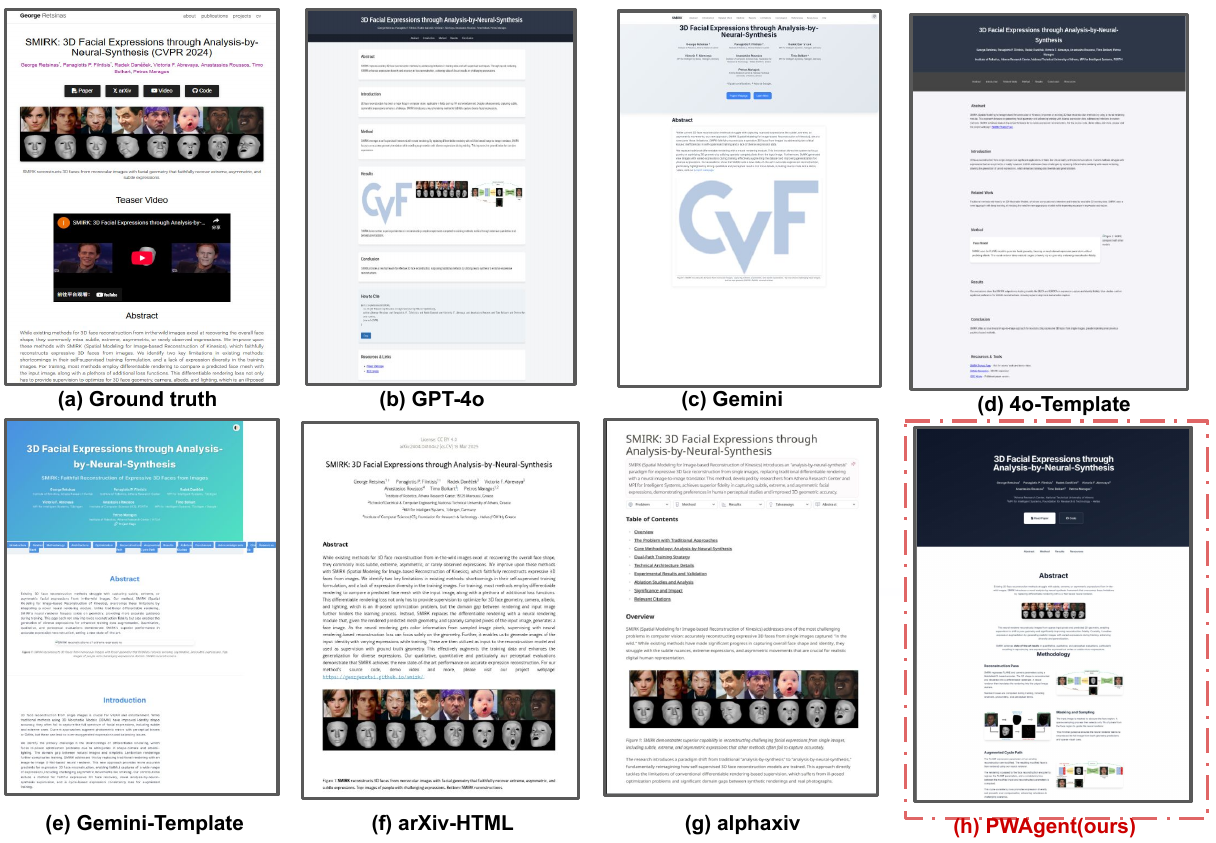}
  \caption{Illustration of website variants for the paper
  ``SMIRK: 3D Facial Expressions through Analysis-by-Neural-Synthesis''\protect\footnotemark generated by different methods.}
  \Description{add_case2}
  \label{fig:add_case2}
  \vspace{-1em}
\end{figure}

\footnotetext{\url{https://georgeretsi.github.io/smirk/}}

\begin{figure}[H]
  \centering
  \includegraphics[width=\textwidth]{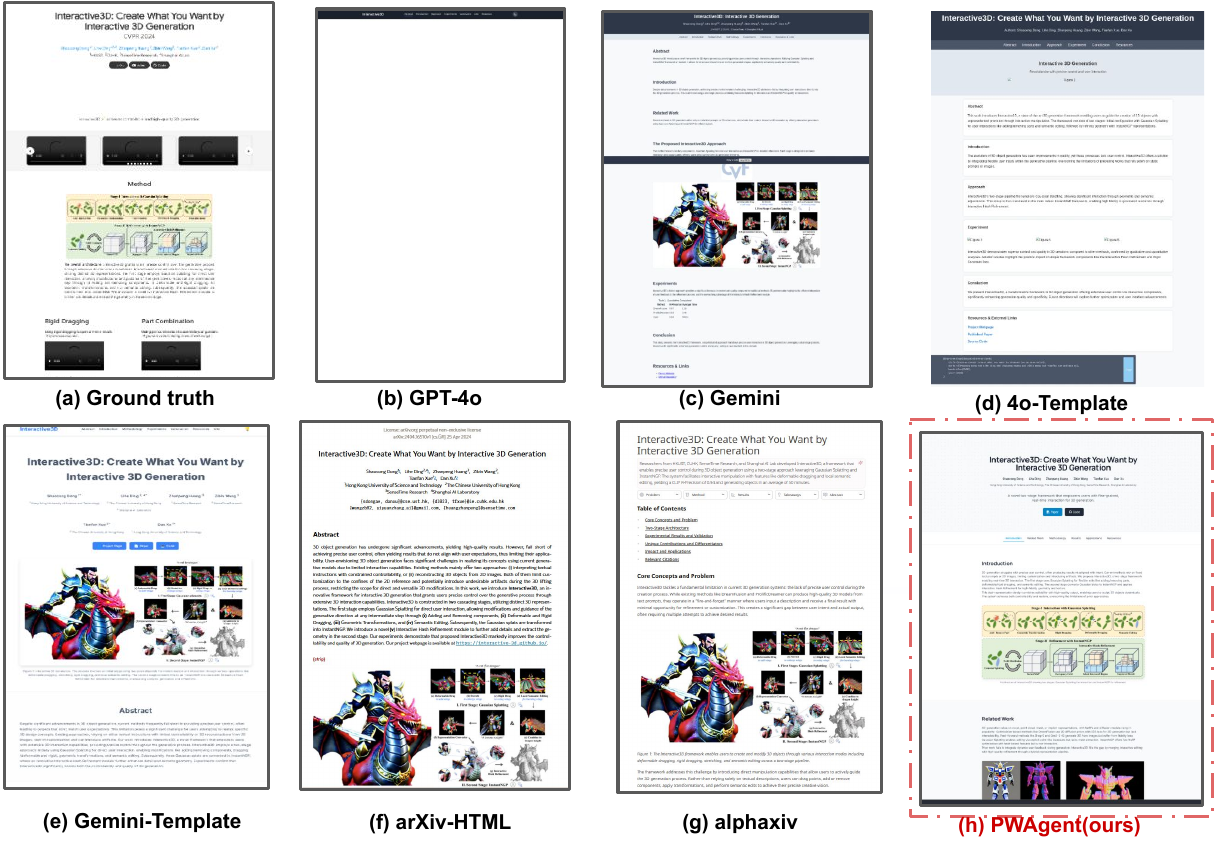}
  \caption{Illustration of website variants for the paper
  ``Interactive3D: Create What You Want by Interactive 3D Generation''\protect\footnotemark generated by different methods.}
  \Description{add_case3}
  \label{fig:add_case3}
  \vspace{-1em}
\end{figure}

\footnotetext{\url{https://interactive-3d.github.io/}}

\begin{figure}[H]
  \centering
  \includegraphics[width=\textwidth]{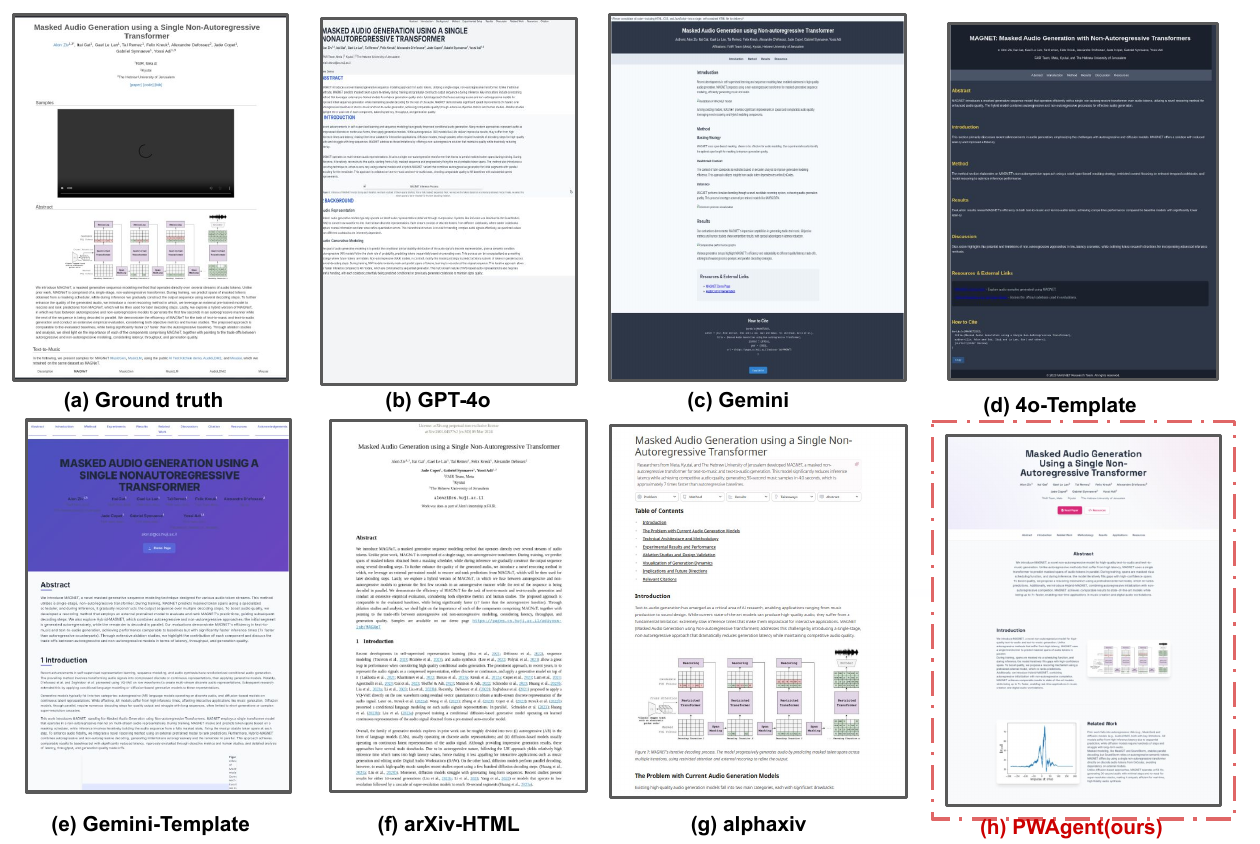}
  \caption{Illustration of website variants for the paper
  ``Masked Audio Generation using a Single Non-Autoregressive Transformer''\protect\footnotemark generated by different methods.}
  \Description{add_case4}
  \label{fig:add_case4}
  \vspace{-1em}
\end{figure}

\footnotetext{\url{https://pages.cs.huji.ac.il/adiyoss-lab/MAGNeT/}}

\begin{figure}[H]
  \centering
  \includegraphics[width=\textwidth]{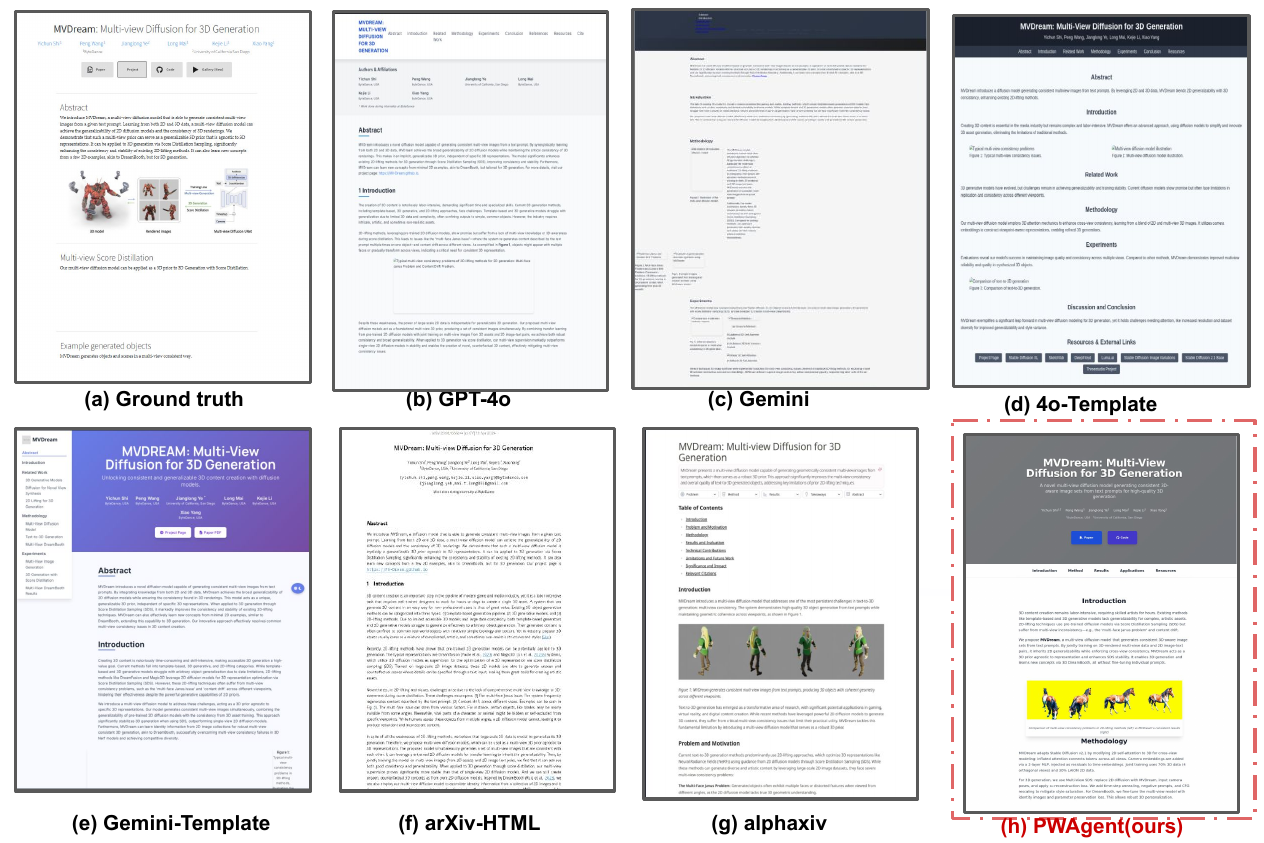}
  \caption{Illustration of website variants for the paper
  ``MVDream: Multi-view Diffusion for 3D Generation''\protect\footnotemark generated by different methods.}
  \Description{add_case5}
  \label{fig:add_case5}
  \vspace{-1em}
\end{figure}

\footnotetext{\url{https://mv-dream.github.io/}}

\end{document}